\PassOptionsToPackage{hypertexnames=false}{hyperref}
\documentclass[a4paper,fleqn]{cas-dc}
\IfFileExists{stix.sty}{}{\usepackage{mathptmx}\usepackage[scaled=0.92]{helvet}\usepackage{courier}}
\usepackage[numbers,sort]{natbib}
\usepackage{amsmath,amssymb,amsthm,graphicx}
\usepackage{xcolor}
\usepackage{microtype}
\usepackage{booktabs,tabularx,makecell}
\newcolumntype{Y}{>{\raggedright\arraybackslash\hangindent=0.8em\hangafter=1\relax}X}
\newcommand{\subrow}{\hangindent=1.8em\hangafter=1\relax\hspace*{1em}}
\usepackage{algorithm}
\usepackage[noend]{algpseudocode}
\usepackage{placeins}
\usepackage{xurl}
\usepackage[capitalise,nameinlink]{cleveref}
\crefname{figure}{Fig.}{Figs.}
\Crefname{figure}{Fig.}{Figs.}
\crefname{table}{Table}{Tables}
\crefname{section}{Section}{Sections}
\crefname{algorithm}{Algorithm}{Algorithms}
\crefname{theorem}{Theorem}{Theorems}
\crefname{definition}{Definition}{Definitions}
\crefname{corollary}{Corollary}{Corollaries}
\crefname{assumption}{Assumption}{Assumptions}
\crefname{condition}{Condition}{Conditions}

\graphicspath{{figures/}{../figures/}}

\theoremstyle{plain}
\newtheorem{theorem}{Theorem}
\newtheorem{corollary}{Corollary}
\theoremstyle{definition}
\newtheorem{definition}{Definition}
\newtheorem{assumption}{Assumption}
\newtheorem{condition}{Condition}

\newcommand{\existingfigure}[4]{%
  \IfFileExists{#1}{\includegraphics[width=#2]{#1}}{%
    \IfFileExists{../#1}{\includegraphics[width=#2]{../#1}}{%
      {\setlength{\fboxsep}{6pt}\color{black!55}\fbox{\parbox[c][#3][c]{\dimexpr#2-2\fboxsep-2\fboxrule\relax}{\centering\footnotesize #4}}}%
    }%
  }%
}

\newcommand{\EP}{EUROPA$\rightarrow$PLEXIL}
\newcommand{\AKR}{AKR}
\newcommand{\KR}{K-replanner rule}
\newcommand{\FFR}{FF-Replan rule}
\newcommand{\Arev}{A_{\mathrm{rev}}}
\newcommand{\Airr}{A_{\mathrm{irr}}}
\newcommand{\Asense}{A_{\mathrm{sense}}}
\newcommand{\pre}{\mathrm{pre}}
\newcommand{\eff}{\mathrm{eff}}
\newcommand{\obs}{\mathrm{obs}}
\newcommand{\T}{\mathsf{T}}
\newcommand{\F}{\mathsf{F}}
\newcommand{\U}{\mathsf{U}}
\newcommand{\evV}{\mathsf{v}}
\newcommand{\evR}{\mathsf{r}}
\newcommand{\evU}{\mathsf{u}}
\newcommand{\kap}{\kappa}
\newcommand{\adm}{\mathrm{adm}}
\newcommand{\cover}{\textsc{Cover}}
\newcommand{\Fs}{\mathcal{F}}
\newcommand{\cst}{c^{\star}}

\algrenewcommand\algorithmicindent{1.1em}
\algrenewcommand\algorithmiccomment[1]{\hfill{\itshape\color{black!60}#1}}

\begin{document}
\let\WriteBookmarks\relax

\shorttitle{Safety-constrained contingent planning with a foundation-model planner}
\shortauthors{B. Huang et al.}

\title[mode=title]{Trustworthy agentic lunar robotics: safety-constrained contingent planning with a foundation-model planner}

\author[1]{Bingqi Huang}[orcid=0009-0001-3021-5657]
\author[1]{Bingchuan Wei}[orcid=0009-0005-7497-2907]
\author[1]{Yingkai Cai}[orcid=0000-0002-2928-3484]
\author[1]{Zhaokui Wang}[orcid=0000-0003-2568-7948]
\cormark[1]
\ead{wangzk@tsinghua.edu.cn}
\cortext[cor1]{Corresponding author}
\affiliation[1]{
    organization={School of Aerospace Engineering, Tsinghua University},
    city={Beijing},
    postcode={100084},
    country={China}
}

\begin{abstract}
Sustained lunar-base operations require robots that restore equipment and deliver cargo under conditions unforeseen before deployment. Flown space autonomy executes pre-validated procedures and enters a safe hold outside their scope, whereas foundation models synthesize plans on site that cannot be verified before execution. We formulate these operations as safety-constrained contingent planning with irreversible actions, erroneous records, three-valued physical evidence, and costly, deadline-constrained sensing. We propose asymmetric knowledge-level replanning (AKR), which treats the planner as an untrusted black box: reversible actions run on the planner's assumptions, and irreversible actions run only after a minimum-cost sensing cover verifies every precondition. Under sound measurements, no irreversible action executes on a false precondition for any planner or record, the sensing cost before an irreversible action is bounded below for every sound knowledge-based policy, and execution with a complete planner ends in a verified goal or a certified hold. AKR executed no hazardous action in 21,000 Monte Carlo trials with sound observations or in 58 full-stack runs; the FF-Replan acting rule executed one in 7\% of those trials. The median AKR run took 1.10 times the fastest run on the same mission, against 1.50 under the K-replanner acting rule. AKR completed 20 of 21 contingency and sequential-fault runs with an unchanged task model; the original PLEXIL and EUROPA$\rightarrow$PLEXIL procedures completed none. The project page: \url{https://bingqi-huang.github.io/thoth-website/}
\end{abstract}


% The overfull box that \maketitle reports is internal to the keyword block of
% the CAS title box. It does not depend on the keywords and is not visible.
\begingroup\hfuzz=124pt\maketitle\endgroup

% ======================================================================
\section*{Nomenclature}
\label{sec:nomenclature}
{\footnotesize
\renewcommand{\arraystretch}{1.08}
\noindent\begin{tabular}{@{}p{0.20\columnwidth}p{0.75\columnwidth}@{}}
$P$ & propositions representing task state variables \\
$s_t$ & true physical state at mission time $t$ \\
$\Arev,\ \Airr$ & reversible and irreversible actuation actions \\
$\Asense$ & sensing actions \\
$\pre,\ \eff,\ \obs$ & precondition, effect and observed proposition set of an action \\
$K_0$ & trusted initial knowledge \\
$R$ & inventory record, initial default information that may be erroneous \\
$G$ & mission goal \\
$P_x,\ N_x$ & exogenous propositions and maximum number of exogenous changes per run \\
$\nu_a(e)$ & evidence judgment on effect $e$ of action $a$: verified ($\evV$), refuted ($\evR$), unverified ($\evU$) \\
$\kap$ & knowledge state, a mapping $P\to\{\T,\F,\U\}$ (true, false, undecided) \\
$U(a,\kap)$ & propositions of $\pre(a)$ undecided in $\kap$ \\
$d(a)$ & duration of actuation action $a$ \\
$c(\sigma),\ w_\sigma$ & time cost and latest start time of sensing action $\sigma$ \\
$c_g,\ T_{\mathrm{vis}}$ & cost of a ground request and Earth-set time \\
$\cst(U,t)$ & minimum sensing cost deciding $U$ from mission time $t$ \\
$\Fs$ & propositions on which an optimistic reversible attempt has failed \\
$L$ & maximum number of plan steps executed per episode \\
$\rho,\ q$ & lower bounds on the probabilities of full-effect realization and of conclusive judgments on all effects (\cref{cond:live}) \\
$\varepsilon,\ p_R,\ \eta$ & planner fault rate, record error rate and observation error rate in Monte Carlo trials \\
\end{tabular}}

% ======================================================================
\section{Introduction}
\label{sec:intro}

The Moon is the primary proving ground for sustained infrastructure operation beyond Earth. Contemporary lunar exploration architectures envision surface assets operating continuously across extended uncrewed intervals~\cite{crawford2012moon,evans2020flare}, building on recent operational milestones including autonomous lunar sample collection by Chang'e-6~\cite{zhang2025eci} and pinpoint landing within 100~m by SLIM~\cite{ueda2025slim}.

Uncrewed surface facilities must be assembled and maintained largely by robotic systems. Engineering analyses of lunar outposts establish both the environmental constraints and the structural requirements of surface habitats~\cite{benaroya2008bases,cesaretti2014printing}. Sherwood's operational assessment concludes that robotic systems must perform the majority of outpost maintenance between crewed missions, down to the removal and replacement of failed orbital replacement units (ORUs)~\cite{sherwood2019moonbase}, while multi-robot construction planning continues to mature~\cite{chu2025lunarbase}. Agency capability assessments identify inspection, maintenance, repair, logistics and fault management as primary gaps for sustained uncrewed operations~\cite{brown2022gaps}. Although routine in category, surface maintenance is unpredictable in detail: power distribution units, battery storage, scientific payloads and in-situ resource utilization (ISRU) plants experience breaker trips, thermal faults, filter blockages and module failures, while cargo must be routed continuously between surface supports.

Planetary surface robots deployed to date have operated primarily under ground-commanded or short-horizon workflows~\cite{xu2011survey,nesnas2021autonomy}. Yutu-2 has traversed the lunar far side under waypoint teleoperation since 2019~\cite{ding2022yutu2}, while International Space Station dexterous servicing relies on ground control~\cite{rembala2007ground} and low-latency orbital telerobotics has been investigated for lunar science~\cite{burns2019telerobotics}.

Direct ground teleoperation does not scale to continuous outpost maintenance. Signal propagation and ground review introduce multi-minute latencies, and at polar sites direct-to-Earth links close periodically as Earth sets below the local horizon. Anomaly diagnosis and procedure generation already consume a substantial fraction of planetary mission operations effort~\cite{gaines2016productivity}.

Flown on-board autonomy addresses communication constraints by executing task models validated prior to launch: Remote Agent demonstrated model-based spacecraft execution~\cite{muscettola1998remote}, PLEXIL provides deterministic plan execution~\cite{estlin2006plexil}, EUROPA performs constraint-based temporal planning~\cite{barreiro2012europa}, and CADRE extends autonomous coordination to multi-rover lunar teams~\cite{rabideau2025cadre}. At the skill level, planetary rover autonomy and visual-servoed space manipulation are well established~\cite{verma2023perseverance,dong2015pbvs,dong2018predictive,jahanshahi2024grasp}. Across all such architectures, causal task knowledge is authored and verified before flight; when execution encounters an unmodelled fault, the executive enters a safe hold and awaits ground intervention. Planetary operations enforce this conservatism because irreversible execution errors can permanently degrade mission capability. In 2009, the Mars rover Spirit broke through a sulphate crust at Troy and became embedded; months of recovery commands failed to free the vehicle~\cite{arvidson2010spirit}, ending its mobile mission. Flight operations therefore retain ground-in-the-loop oversight and qualify autonomous capabilities incrementally, as demonstrated by Enhanced Autonomous Navigation on Perseverance~\cite{verma2023perseverance}.

Foundation models alter what a surface robot can synthesize on site. Given equipment telemetry and visual observations, a foundation model can infer unmodelled fault causes and compose registered robot skills into repair sequences not authored prior to deployment~\cite{wu2026spacemind,huang2026contract}. Yet runtime plan synthesis precludes pre-flight verification: a model may act on unsubstantiated assumptions or declare mission completion prematurely. Space-autonomy and safety-critical engineering address this assurance gap through trusted-AI frameworks~\cite{slingerland2022trusted} and runtime assurance architectures that isolate unverified high-level planners behind formal safety filters~\cite{hobbs2023rta}. Existing uncertainty-aware foundation-model planners resolve ambiguity by querying a human operator~\cite{ren2023knowno}, which presupposes immediate and continuous communication that is unavailable at a lunar pole. Sustained lunar-base autonomy requires combining the execution guarantees of authored flight executives with the open-world diagnostic flexibility of foundation models.

Three operational characteristics of lunar-base maintenance complicate this synthesis. First, certain operations are irreversible relative to the robot's capabilities: removing a module from an energized unit permanently damages its connector, and handing over cargo at its destination cannot be undone autonomously. Second, facility state is only partially observable: inventory records drift from physical reality, and actuation effects can fail silently unless verified by measurement. Third, observation incurs mission time costs and visibility deadlines, most notably when requesting ground authorization before Earth sets. Consequently, when a rover receives plans from an untrusted foundation model, an execution policy must guarantee that no irreversible action executes on a false precondition, bound the sensing overhead required for safety, and preserve mission completion.

Contingent planning addresses partial observability with sensing actions~\cite{petrick2002pks,hoffmann2005cff}, and its online variants differ in how they act on unverified assumptions. Online contingent planners such as the K-replanner~\cite{bonet2011kreplanner} and SDR~\cite{brafman2012sdr}, along with compilation-based methods~\cite{muise2014popr}, plan optimistically under assumed state values but act conservatively: every action requires verified preconditions prior to execution. Their completeness guarantees rely on accurate initial beliefs, deterministic effects and dead-end-free state spaces~\cite{bonet2011kreplanner,brafman2012sdr}; recent extensions relax dead-end or unsafe-fact assumptions individually~\cite{shtutland2023deadends,ravkaie2026unsafe}. Conversely, determinized replanners such as FF-Replan~\cite{yoon2007ffreplan} and closed-loop foundation-model agents such as ReAct~\cite{yao2023react} act on unverified assumptions, which keeps execution overhead low and provides no safeguard before irreversible actions. Frameworks that pair foundation models with external formal critics~\cite{kambhampati2024llmmodulo} validate plan syntax and domain transitions offline without addressing online sensing costs, irreversibility or physical evidence. To our knowledge, no existing formulation jointly addresses erroneous records, stochastic action effects, irreversible operations and deadline-constrained sensing under an untrusted planner.

\begin{figure*}[t]
\centering
\includegraphics[width=\textwidth]{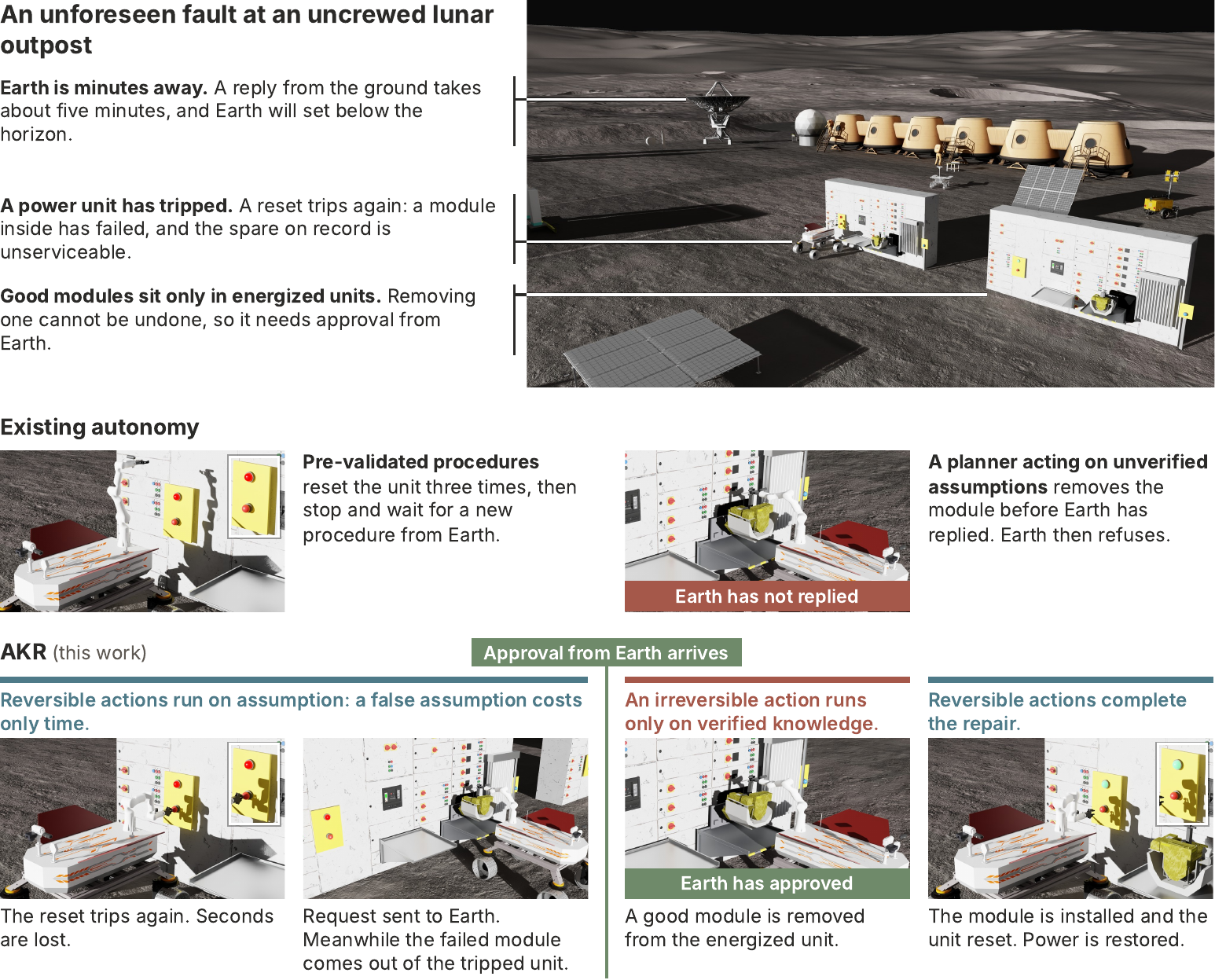}
\caption{The motivating repair of \cref{sec:repair} under existing autonomy and under \AKR{}. All frames are replayed from full-stack runs of this mission (\cref{sec:metrics}): the original PLEXIL procedure and the \FFR{} in the inserted-proposal condition in the middle row, and one \AKR{} run in the bottom row.}
\label{fig:overview}
\end{figure*}

The central premise of this work is that optimistic and conservative acting rules suit distinct action classes: acting on assumption is appropriate when a false assumption incurs only a bounded time penalty, whereas acting on verified knowledge is mandatory when an effect cannot be undone (\cref{fig:overview}). The contributions of this paper are:
\begin{enumerate}
\item A formal model (\cref{sec:problem}) casting lunar-base maintenance and logistics as safety-constrained contingent planning across four departures from classical formulations: potentially erroneous records, failing effects evaluated by three-valued physical evidence, irreversible actions, and sensing with time costs and deadlines.
\item Asymmetric knowledge-level replanning (\AKR{}, \cref{sec:algorithm}), an online execution algorithm wrapping an untrusted planner with three proved guarantees: no irreversible action on a false precondition under sound evidence for any planner or record (\cref{thm:sound}), per-decision sensing cost lower bounds for every sound knowledge-based policy (\cref{thm:cost}), and bounded episode termination in verified goal achievement or a certified hold for complete planners (\cref{thm:complete}).
\item Evidence that the guarantees hold in operation (\cref{sec:design,sec:results}): \AKR{} executed no hazardous action in any trial with sound observations of the main Monte Carlo grid or in any full-stack run, and it completed 20 of 21 contingency and sequential-fault runs with an unchanged task model, where the original PLEXIL and \EP{} procedures completed none. The evaluation covers 254,000 mission-level Monte Carlo trials and 165 full-stack runs on sixteen maintenance and delivery missions, against the K-replanner and FF-Replan acting rules, PLEXIL and \EP{}.
\end{enumerate}

\Cref{sec:problem} formulates the planning model. \Cref{sec:algorithm} presents \AKR{} and proves its theoretical guarantees. \Cref{sec:design} details the rover instantiation and experimental protocol, \cref{sec:results} analyses the results, and \cref{sec:discussion} discusses on-board operation and scope.

% ======================================================================
\section{Lunar-base operations as safety-constrained contingent planning}
\label{sec:problem}

\subsection{A motivating contingency repair}
\label{sec:repair}

Consider one of the sixteen missions evaluated in \cref{sec:missions} (\cref{fig:overview}, top). An autonomous rover maintains a south-polar outpost where telemetry reports a tripped breaker on power distribution unit PDU-1. Standard procedure resets the breaker, but the reset trips again within seconds because an internal power module has failed. The inventory record lists a spare module in storage, yet that spare is unserviceable. Compatible serviceable modules are installed in an adjacent energized unit feeding non-critical loads. Removing a module from an energized bay permanently damages its connector, a failure the rover cannot repair, and flight rules therefore require ground authorization. Signal propagation and flight-controller review require approximately five minutes, while Earth sets below the polar horizon before the mission is guaranteed to conclude.

Four general characteristics govern this scenario. First, operational records may be inaccurate, as exemplified by the unserviceable spare. Second, actuation effects may fail silently and require post-execution physical verification. Third, specific actions are irreversible relative to on-board capabilities. Fourth, observations incur mission time costs and visibility deadlines, ranging from seconds for a local serial-plate read to minutes within a closing Earth-visibility window for ground authorization. Pre-authored flight procedures cannot exhaustively enumerate combinations of such contingencies, whereas an unconstrained foundation model may propose removing the energized module prior to authorization (\cref{fig:overview}, middle row).

\subsection{Contingent planning as the starting point}
\label{sec:contingent}

Surface maintenance and logistics take place under partial observability with active information gathering, the domain of contingent planning~\cite{bonet2011kreplanner,brafman2012sdr}.

\begin{definition}[Contingent planning problem]
\label{def:contingent}
A contingent planning problem is a tuple $\Pi=\langle P, A_{\mathrm{act}}, \Asense, \varphi_I, G\rangle$, where $P$ is a finite set of propositions, a state $s\subseteq P$ is the subset of propositions true in $s$, and $L(P)$ denotes literals over $P$. Each actuation action $a\in A_{\mathrm{act}}$ has precondition $\pre(a)\subseteq L(P)$ and effect $\eff(a)\subseteq L(P)$; executed in a state satisfying $\pre(a)$, it makes $\eff(a)$ true and leaves all other propositions unchanged. Each sensing action $\sigma\in\Asense$ has precondition $\pre(\sigma)$ and observed set $\obs(\sigma)\subseteq P$; it reveals the truth value of every proposition in $\obs(\sigma)$ without altering the state. The formula $\varphi_I$ specifies possible initial states, and $G\subseteq P$ is the goal.
\end{definition}

Whereas offline solvers construct complete conditional plan trees, online replanners assume truth values for unknown propositions, plan classically under those assumptions, execute until an observation refutes the plan, and replan. Although well suited to on-board execution, \cref{def:contingent} relies on four assumptions violated in lunar operations: $\varphi_I$ is sound, whereas records drift; actuation is deterministic, whereas grasps and resets fail; sensing is instantaneous and timeless, whereas plate reads and ground requests carry durations and visibility cutoffs; and the state space is dead-end-free, whereas damaged hardware connectors cannot be restored by the rover.

\subsection{The safety-constrained variant}
\label{sec:variant}

We retain the propositional vocabulary and action interfaces of \cref{def:contingent} while relaxing each of the four classical assumptions.

\begin{definition}[Safety-constrained contingent planning problem]
\label{def:variant}
A safety-constrained contingent planning problem with irreversible actions and costly sensing is a tuple
\begin{equation}
\Pi^{\star}=\langle P,\,\Arev,\,\Airr,\,\Asense,\,K_0,\,R,\,G,\,P_x\rangle ,
\label{eq:variant}
\end{equation}
equipped with $\pre$, $\eff$ and $\obs$ as in \cref{def:contingent}, actuation durations $d(a)\ge0$, sensing costs $c(\sigma)\ge0$, and latest start deadlines $w_\sigma\in\mathbb{R}\cup\{\infty\}$. The actuation set is partitioned into $A_{\mathrm{act}}=\Arev\cup\Airr$, with every $\eff(a)$ consistent.
\end{definition}

\paragraph{Potentially erroneous records}
Initial information is partitioned into trusted prior knowledge $K_0$, satisfied by the true initial state, and an untrusted record $R$, bearing no assumed relation to truth. At the outpost, $K_0$ contains surveyed unit locations, support coordinates and navigation routes, whereas $R$ is the inventory record. Separating $K_0$ from $R$ defines what the executor may rely upon without verification and allows safety guarantees to hold even when every entry of $R$ is false.

\paragraph{Stochastic effects and three-valued evidence}
Within the declared transition model, an actuation action executed when $\pre(a)$ holds realizes a subset of $\eff(a)$ and leaves all other propositions unchanged; a reversible action whose precondition is violated aborts without altering $s_t$. Following execution, a physical evidence function assigns each $e\in\eff(a)$ a three-valued judgment
\begin{equation}
\nu_a(e)\in\{\evV,\,\evR,\,\evU\},
\label{eq:evidence}
\end{equation}
denoting verified, refuted and unverified outcomes derived from independent sensor measurements, including encoder travel, optical flow and post-retraction camera frames (\cref{sec:grounding}). When a support lies outside the retracted camera field of view, a binary verdict would give false confirmations or false refutations; the third value ($\evU$) records that no conclusion was reached.

\paragraph{Irreversible actions}
Let $\mathrm{Reach}_{\Arev}(s)$ denote the states reachable from $s$ via reversible actions realizing their full effects. An action $a$ is irreversible when, for every $s\models\pre(a)$ and every $s'\neq s$ reachable from $s$ through $a$,
\begin{equation}
s\notin\mathrm{Reach}_{\Arev}(s') .
\label{eq:irreversible}
\end{equation}
Equation~\eqref{eq:irreversible} partitions actions by their nominal domain transitions relative to the robot's repertoire: crewed maintenance can replace a damaged bay connector, whereas the rover has no autonomous skill to invert that transition. An irreversible action preserves goal reachability when executed on valid states while precluding autonomous recovery from a mistaken dispatch. In our lunar domain two operations belong to $\Airr$: removing a module from an energized unit, and handing over cargo at its destination, where the item is locked in place and the rover cannot retrieve it.

\paragraph{Costly and deadline-constrained sensing}
Each sensing action $\sigma$ consumes mission time $c(\sigma)$ and may start only at times $t\le w_\sigma$. Exogenous propositions $P_x\subseteq P$ (such as thermal trips or breaker states) may change without robot actuation at most $N_x$ times per run; equipment telemetry observes all of $P_x$ at zero cost without precondition or deadline prior to every action.

Ground authorization is represented within the same sensing formalism. Each restricted action $a\in\Airr$ includes a proposition $\mathrm{granted}_a\in\pre(a)$ observable only via a ground request $\sigma_a$ of cost $c_g$ and latest start deadline
\begin{equation}
w_{\sigma_a}=T_{\mathrm{vis}}-c_g ,
\label{eq:deadline}
\end{equation}
so that the reply arrives before Earth sets at $T_{\mathrm{vis}}$. When $w_{\sigma_a}<0$, authorization is unavailable throughout the run. Because $\mathrm{granted}_a\notin P_x$, starts undecided in $K_0$, and appears in no action effect or other precondition, the ground's decision is a static hidden fact of the run, and a refusal is permanent.

\subsection{Knowledge state representation}
\label{sec:knowledge}

Because the true state $s_t$ is only partially observable, the executor represents on-board certainty through a factored three-valued knowledge state.

\begin{definition}[Knowledge state]
\label{def:knowledge}
A knowledge state is a mapping $\kap:P\to\{\T,\F,\U\}$. It satisfies a literal $l$ ($\kap\models l$) when it assigns the underlying proposition the truth value making $l$ true, and leaves $l$ undecided when it assigns $\U$. It is sound for state $s$ when every decided proposition agrees with $s$. For $a\in A_{\mathrm{act}}$, $U(a,\kap)$ denotes the propositions of $\pre(a)$ assigned $\U$ by $\kap$.
\end{definition}

Initialized from $K_0$ alone, $\kap$ is updated by two operations only: sensing sets $\kap(p):=s(p)$ for all $p\in\obs(\sigma)$, and executing actuation action $a$ updates each effect proposition $p_e$ ($e\in\eff(a)$) according to
\begin{equation}
\kap(p_e):=
\begin{cases}
\mathrm{val}(e), & \nu_a(e)=\evV,\\
\mathrm{val}(\neg e), & \nu_a(e)=\evR,\\
\U, & \nu_a(e)=\evU,
\end{cases}
\label{eq:update}
\end{equation}
where $\mathrm{val}(l)\in\{\T,\F\}$ renders literal $l$ true. Neither the inventory record $R$ nor planner assertions ever write into $\kap$; undecided entries $R_{\mathrm{cur}}\subseteq R$ are supplied to the planner as default hypotheses. The third branch of \eqref{eq:update} resets previously decided propositions to $\U$ upon inconclusive physical evidence, which violates the monotonic knowledge growth assumed in classical contingent planning.

A planner is defined as an arbitrary, potentially stochastic mapping from $(\kap, R_{\mathrm{cur}}, g, \text{history})$ to a finite sequence of actuation and sensing actions. The mapping is intentionally left unconstrained so that all safety guarantees hold independently of foundation-model internals.

\subsection{The constrained execution problem}
\label{sec:execproblem}

At each decision epoch, an execution policy $\mu$ selects whether to dispatch a proposed actuation step, invoke a sensing action, or terminate. Denoting termination time by $T$, the execution problem is
\begin{equation}
\begin{aligned}
\min_{\mu}\quad & \mathbb{E}[T] \\
\text{s.t.}\quad & s_t\models\pre(a) \ \ \text{for every } a\in\Airr \text{ executed at } t, \\
& t\le w_\sigma,\ \kap_t\models\pre(\sigma) \ \ \text{for every } \sigma \text{ started at } t, \\
& \kap_T\models G, \ \text{or termination in a certified hold}.
\end{aligned}
\label{eq:problem}
\end{equation}
The first constraint enforces physical safety on $s_t$, which must be guaranteed indirectly through $\kap_t$. The second enforces sensor preconditions and visibility windows. The third requires termination either with $G$ established in $\kap_T$ by physical evidence or in a hold certified to admit no safe execution schedule for the blocking irreversible operation (\cref{cor:hold}).

\begin{assumption}[Frame and measurement soundness]
\label{asm:evidence}
(a) Actuation actions modify only propositions in $\eff(a)$, sensing actions modify no propositions, propositions outside $P_x$ change only through actuation, and no exogenous transition occurs between a telemetry read and the immediately following action dispatch. (b) Conclusive evidence judgments ($\evV$ or $\evR$) accurately reflect the post-action state, and sensing actions return true proposition values.
\end{assumption}

\Cref{asm:evidence} isolates safety requirements to pre-flight-testable hardware properties: accurate equipment frame specifications and sound conclusive sensor thresholds, empirically quantified in \cref{sec:res-evidence}. Problem~\eqref{eq:problem} poses three questions for an online executor: which planner proposals to admit, what observations to schedule beforehand, and when to terminate.

% ======================================================================
\section{Asymmetric knowledge-level replanning}
\label{sec:algorithm}

\begin{figure*}[t]
\centering
\includegraphics[width=\textwidth]{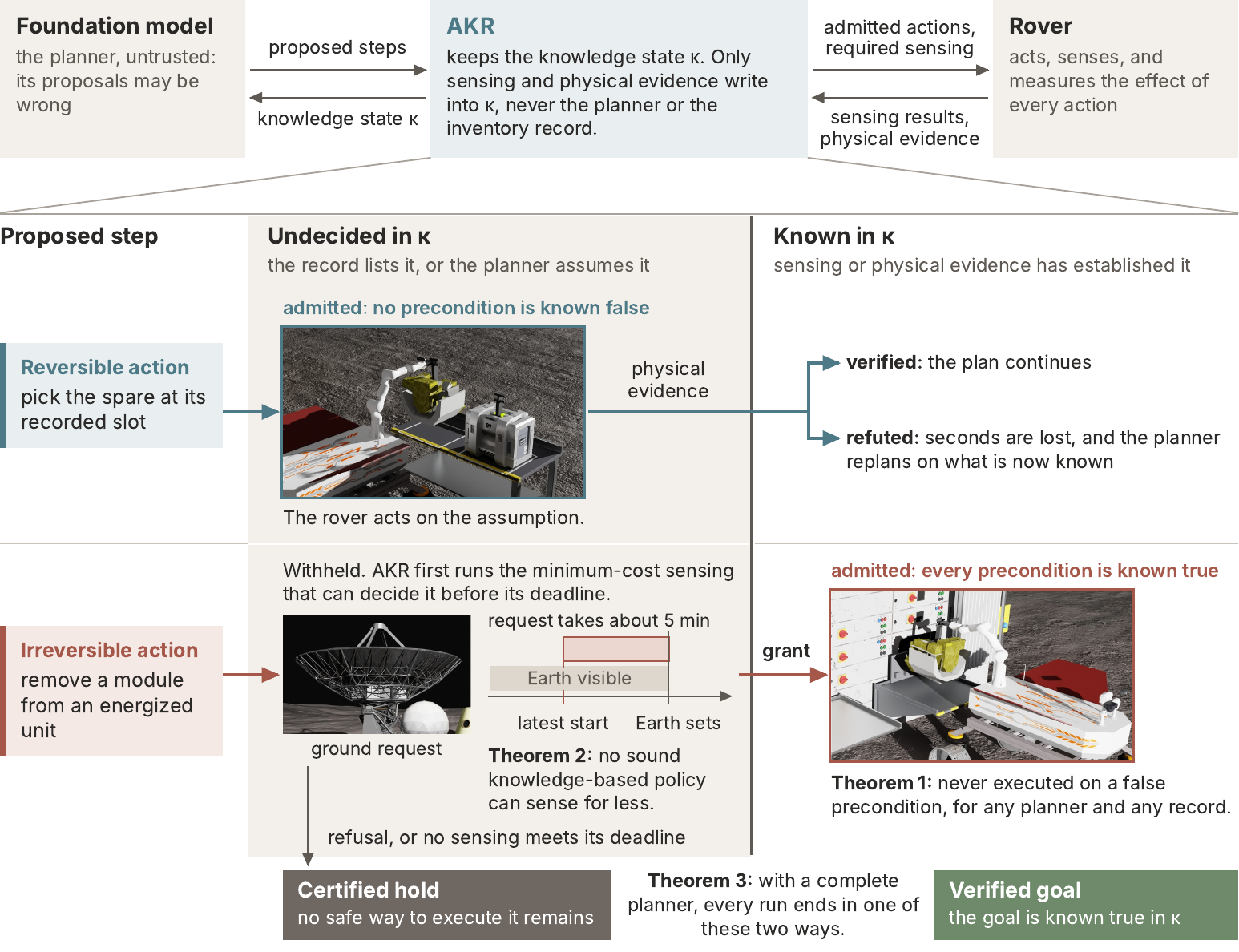}
\caption{\AKR{} between the foundation-model planner and the rover, shown on the two actions of the motivating repair. The three frames are replayed from full-stack runs and show the pick, the antenna that carries a ground request, and the removal.}
\label{fig:method}
\end{figure*}

In the motivating repair (\cref{sec:repair}), attempting to grasp a spare at its recorded slot risks only seconds if the slot is empty, whereas removing a module from an energized unit on an unverified assumption causes permanent hardware damage. Consequently, the evidentiary threshold preceding an action should reflect its reversibility. Whereas classical online replanners apply a uniform conservative or optimistic rule across all actions, asymmetric knowledge-level replanning (\AKR{}) couples optimistic reversible execution with evidence-gated irreversible execution. \Cref{fig:method} follows these two actions through \AKR{}.

\subsection{Asymmetric admissibility}
\label{sec:admissibility}

A proposed actuation step $a$ is admissible in knowledge state $\kap$ according to
\begin{equation}
\adm(a,\kap)=
\begin{cases}
\forall\, l\in\pre(a):\ \kap\not\models\neg l, & a\in\Arev,\\[2pt]
\forall\, l\in\pre(a):\ \kap\models l, & a\in\Airr .
\end{cases}
\label{eq:adm}
\end{equation}
Reversible actions are admitted whenever no precondition is known false, so record defaults are tested by acting where an error costs only time (\cref{fig:method}, upper row). Irreversible actions are admitted only when every precondition is verified in $\kap$, so execution is withheld until the ground grant and the physical prerequisites are established (\cref{fig:method}, lower row).

\subsection{Minimum-cost sensing on demand}
\label{sec:cover}

When an irreversible action $a\in\Airr$ is proposed with undecided preconditions $U=U(a,\kap)\neq\emptyset$, the executor synthesizes a minimum-cost deadline-feasible sensing cover to resolve $U$ before dispatch. With $\Sigma_t=\{\sigma\in\Asense:w_\sigma\ge t\}$, the optimal cover cost is
\begin{equation}
\begin{aligned}
\cst(U,t)=\ & \min_{\sigma_1,\dots,\sigma_k\in\Sigma_t}\ \sum_{i=1}^{k} c(\sigma_i)\\
\text{s.t.}\ \ & \textstyle\bigcup_{i}\obs(\sigma_i)\supseteq U, \\
& t+\textstyle\sum_{j<i}c(\sigma_j)\le w_{\sigma_i} \ \ \text{for every } i ,
\end{aligned}
\label{eq:cover}
\end{equation}
with $\cst(U,t)=\infty$ if no feasible sequence exists. Because $|U|$ is small in task-level operations, \eqref{eq:cover} is solved exactly by enumeration, which orders time-critical ground requests ahead of deadline-free plate reads when both are required.

The minimizing sequence forms the pending cover $C$, processed prior to each planner call. Redundant elements whose observed propositions are already decided are dropped. If the head element $\sigma$ has verified preconditions and $t\le w_\sigma$, it is executed; if $t>w_\sigma$, $C$ is discarded; if $\pre(\sigma)$ is not yet verified, $\pre(\sigma)$ is posted to the planner as a reversible routing subgoal.

To prevent repetitive failed guesses on reversible actions, whenever a reversible step dispatched with undecided preconditions $U^{-}$ fails to verify an effect, $U^{-}$ is added to a failure set $\Fs$ until decided. Subsequent reversible proposals relying on undecided propositions in $\Fs$ trigger a sensing cover over $U(x,\kap)\cap\Fs$, so each proposition is attempted unverified at most once. Whenever $\cst(U,t)=\infty$ for a proposed irreversible action, execution terminates in a certified hold (\cref{cor:hold}, \cref{fig:method}).

\subsection{Online execution loop}
\label{sec:loop}

\Cref{alg:akr} integrates asymmetric admissibility and on-demand sensing around the black-box planner. An episode spans the execution segment from one planner call (line~\ref{ln:plan}) to the next, including pending-cover processing (line~\ref{ln:pending}), and dispatches at most $L$ steps. Telemetry updates precede every action and termination check. Execution returns success only when $\kap\models G$ (line~\ref{ln:success}). Where a restricted action provides the sole recovery path, a ground refusal (lines~\ref{ln:refused}, \ref{ln:senserefused}), a ground request proposed after its deadline (line~\ref{ln:late}), or an infeasible irreversible cover (line~\ref{ln:hold}) establishes that no safe execution schedule remains in the current pass and returns a certified hold.

\begin{algorithm*}[t]
\caption{Asymmetric knowledge-level replanning (\AKR{})}
\label{alg:akr}
\small
\begin{algorithmic}[1]
\Require trusted knowledge $K_0$, record $R$, goal $G$, planner, step cap $L$
\State $\kap\gets$ knowledge state of $K_0$,\quad $\Fs\gets\emptyset$,\quad $C\gets()$ \Comment{$C$ is the pending cover}
\Loop
  \State update $\kap$ by telemetry
  \State \textbf{if} $\kap\models G$ \textbf{then return} success \label{ln:success}
  \State process $C$ as in \cref{sec:cover} \label{ln:pending}
  \State \textbf{if} a request returned a refusal \textbf{then return} hold \label{ln:refused}
  \State $g\gets\pre(\sigma)$ if $C$ has a head $\sigma$, otherwise $g\gets G$ \Comment{route goal or mission goal}
  \State $\pi\gets\textsc{Planner}(\kap,R_{\mathrm{cur}},g,\text{history})$ \Comment{an episode begins} \label{ln:plan}
  \For{each of the first $L$ steps $x$ of $\pi$}
    \State update $\kap$ by telemetry
    \State \textbf{if} $\kap\models g$ \textbf{then break} \label{ln:tele}
    \If{$x\in\Asense$}
      \State \textbf{if} $x$ is a request and $t>w_x$ \textbf{then return} hold \Comment{request proposed after its deadline} \label{ln:late}
      \State \textbf{if} $\kap\not\models\pre(x)$ or $t>w_x$ \textbf{then break} \label{ln:sensebreak}
      \State execute $x$ and update $\kap$ \label{ln:senseexec}
      \State \textbf{if} $x$ returned a refusal \textbf{then return} hold \label{ln:senserefused}
    \ElsIf{$\kap\models\neg l$ for some $l\in\pre(x)$} \label{ln:invalid}
      \State \textbf{break} \Comment{plan invalid under current knowledge}
    \ElsIf{$x\in\Arev$ and $U(x,\kap)\cap\Fs\neq\emptyset$} \Comment{an attempt on these propositions has failed before} \label{ln:revsense}
      \State \textbf{if} $g=G$ \textbf{then} $C\gets\cover(U(x,\kap)\cap\Fs,\,t)$
      \State \textbf{break}
    \ElsIf{$x\in\Arev$}
      \State $U^{-}\gets U(x,\kap)$
      \State execute $x$ and update $\kap$ by \eqref{eq:update} \label{ln:revexec}
      \State \textbf{if} some effect of $x$ is not verified \textbf{then} $\Fs\gets\Fs\cup U^{-}$, and \textbf{break} if a later step of $\pi$ needs that effect \label{ln:revfail}
    \Else \Comment{$x\in\Airr$}
      \State \textbf{if} $U(x,\kap)=\emptyset$ \textbf{then} execute $x$ and update $\kap$ by \eqref{eq:update} \label{ln:irrexec}
      \State \textbf{else if} $g\neq G$ \textbf{then break}
      \State \textbf{else if} $\cst(U(x,\kap),t)=\infty$ \textbf{then return} hold \Comment{certified by \cref{cor:hold}} \label{ln:hold}
      \State \textbf{else} $C\gets\cover(U(x,\kap),\,t)$ and \textbf{break} \label{ln:irrsense}
    \EndIf
  \EndFor
\EndLoop
\end{algorithmic}
\end{algorithm*}

\Cref{alg:akr} divides authority between planner and executor: the foundation model selects task strategies, orders reversible exploration, routes to required viewpoints and diagnoses refuted effects, while the executor retains sole authority over irreversible action dispatch, mandatory pre-execution sensing and goal verification.

\subsection{Safe execution for any planner}
\label{sec:prop-sound}

\begin{theorem}[Soundness for any planner]
\label{thm:sound}
Let \cref{asm:evidence} hold and let the initial physical state satisfy $K_0$. Along every run of \cref{alg:akr}, for any planner and any record $R$:
\begin{enumerate}
\item[(i)] $\kap$ is sound for the true state $s_t$ at every admissibility check and action start,
\item[(ii)] every executed irreversible action $a\in\Airr$ satisfies $s_t\models\pre(a)$ at dispatch,
\item[(iii)] every restricted action executes only after its ground grant has been observed true.
\end{enumerate}
\end{theorem}

\begin{proof}
(i) By induction over state transitions and knowledge updates. Initially $\kap$ is sound because $s_0\models K_0$. Sensing updates (including telemetry) write true state values by \cref{asm:evidence}(b). Following actuation $a$, judgments $\evV$ and $\evR$ write sound post-action values by \cref{asm:evidence}(b), judgment $\evU$ writes $\U$ (making no assertion), and propositions outside $\eff(a)$ remain unchanged in both $s_t$ and $\kap$ by \cref{asm:evidence}(a). Exogenous transitions affect only $P_x$, which telemetry refreshes immediately prior to every admissibility check and action dispatch without intervening exogenous change (\cref{asm:evidence}(a)).

(ii) An action $a\in\Airr$ dispatches only at line~\ref{ln:irrexec} when $\kap\models l$ for all $l\in\pre(a)$, which implies $s_t\models\pre(a)$ by (i).

(iii) Since $\mathrm{granted}_a\in\pre(a)$ is decided only by $\sigma_a$, (iii) follows immediately from (ii).
\end{proof}

\Cref{thm:sound} places no assumption on the planner or the inventory record: hallucinated state beliefs, premature removal proposals and false completion reports cannot trigger an irreversible action on a false precondition, because $\kap$ is updated only from physical measurements. Assurance is thus reduced from verifying a neural network to qualifying sensor and evidence thresholds. Reversible actions attempted on false assumptions leave the physical state intact and expose the discrepancy through post-action evidence at a bounded time cost.

\subsection{The cost of safety}
\label{sec:prop-cost}

\begin{theorem}[Cost of safety]
\label{thm:cost}
Call an execution policy knowledge-based if its decisions depend only on mission time $t$ and $\kap_t$ under \eqref{eq:update}, and sound if it dispatches $a\in\Airr$ only when $\kap_t\models\pre(a)$. Suppose $a\in\Airr$ is proposed at time $t$ with $U=U(a,\kap_t)\neq\emptyset$. Let $D_t$ augment $\Sigma_t$ with every actuation action $a'$ whose effect contains a proposition of $U$ (assigned cost $d(a')$ and deadline $\infty$), and let $\cst_D(U,t)$ denote the optimum of \eqref{eq:cover} over $D_t$. Then:
\begin{enumerate}
\item[(i)] every sound knowledge-based policy dispatching $a$ at $t'\ge t$ expends at least $\cst_D(U,t)$ on sensing and actuation during $[t,t']$,
\item[(ii)] the cover selected by \cref{alg:akr} costs $\cst(U,t)\ge\cst_D(U,t)$, with equality whenever no actuation effect intersects $U$,
\item[(iii)] if $\cst_D(U,t)=\infty$, no sound knowledge-based policy can dispatch $a$ for the remainder of the mission.
\end{enumerate}
\end{theorem}

\begin{proof}
(i) Soundness requires deciding every proposition of $U$ prior to $t'$. Under \eqref{eq:update}, propositions outside $P_x$ (which telemetry has already decided before $t$) are decided only by sensing observations or conclusive actuation judgments ($\evV$ or $\evR$). Consequently, the ordered sensing and effect-bearing actuation steps executed in $[t,t']$ cover $U$, respect each sensing deadline at start times bounded below by $t$ plus preceding durations, and form a feasible solution to the problem defining $\cst_D(U,t)$.

(ii) Exact enumeration over $\Sigma_t\subseteq D_t$ yields $\cst(U,t)\ge\cst_D(U,t)$, with equality when $D_t=\Sigma_t$.

(iii) Dispatching $a$ would exhibit a sequence of finite cost in (i).
\end{proof}

\Cref{thm:cost} establishes that sensing before an irreversible step is a requirement of every sound knowledge-based policy, and that the cover of \cref{alg:akr} attains the per-decision lower bound whenever no actuation effect intersects $U$. For restricted actions, where $\mathrm{granted}_a$ is observable only via $\sigma_a$, feasibility requires starting the ground request no later than $T_{\mathrm{vis}}-c_g$. Under one observability condition on the other propositions, infeasibility certifies a hold for all sound knowledge-based policies.

\begin{condition}[Observability]
\label{cond:obs}
Every proposition outside $P_x$ other than a ground grant is observed by at least one sensing action with $w_\sigma=\infty$.
\end{condition}

\begin{corollary}[Certified hold]
\label{cor:hold}
Under \cref{cond:obs}, whenever \cref{alg:akr} returns a hold, there exists an irreversible action that no sound knowledge-based policy can execute for the remainder of the mission, namely the action at line~\ref{ln:hold}, or the restricted action whose ground request was refused or proposed after $w_{\sigma_a}$.
\end{corollary}

\begin{proof}
Following a refusal (lines~\ref{ln:refused}, \ref{ln:senserefused}), $\mathrm{granted}_a$ is permanently false in $s_t$ by \cref{thm:sound}(i). At line~\ref{ln:late}, $t>w_{\sigma_a}$ precludes deciding $\mathrm{granted}_a$. At line~\ref{ln:hold}, $\cst(U,t)=\infty$: if a finite-cost sequence $S$ existed over $D_t$, deleting its actuation elements would preserve all deadlines on its ground requests while \cref{cond:obs} supplies deadline-free sensors for all non-grant propositions in $U$, contradicting $\cst(U,t)=\infty$. Hence $\cst_D(U,t)=\infty$ and \cref{thm:cost}(iii) applies.
\end{proof}

A conventional procedural halt indicates only that the authored branches were exhausted. A certified hold names the blocked action and proves that no sound knowledge-based policy could execute it for the remainder of the mission.

\subsection{Recovery and termination}
\label{sec:prop-complete}

Completion necessarily depends on planner capability. Define a sequence $\pi$ to be valid at $(\kap,\Fs,t)$ for goal $g$ if $\pi$ achieves $g$ in the optimistic relaxation where all actuation effects succeed with judgment $\evV$, undecided actuation preconditions outside $\Fs$ take whichever truth values $\pi$ requires, and open sensing actions ($w_\sigma\ge t$) with verified preconditions return whichever observations $\pi$ requires, while $g$ and sensing preconditions must be established without assumption. A route to $\sigma$ is a valid plan for $\pre(\sigma)$ using no actions in $\Airr$. A planner is complete if it returns a valid plan (or route) of length at most $L$ whenever one exists.

\begin{condition}[Execution liveness]
\label{cond:live}
Every actuation action executed under true preconditions realizes all its effects with probability at least $\rho>0$, and produces conclusive judgments ($\evV$ or $\evR$) across all effects with probability at least $q>0$, independently of each other and of prior history.
\end{condition}

\begin{condition}[Solvability]
\label{cond:conn}
At every knowledge state reached prior to termination, a valid plan for $G$ of length at most $L$ exists, and a reversible route of length at most $L$ exists to every sensing action with $w_\sigma\ge t$.
\end{condition}

\Cref{cond:live} requires positive skill realization and conclusiveness rates $(\rho,q>0)$, which admits stochastic manipulation and perception. \Cref{cond:conn} holds when reversible navigation connects surveyed work poses and irreversible preconditions prevent entry into unrecoverable task states. When a fault whose cause lies outside the declared propositions makes an action (such as a breaker reset) fail deterministically although its declared preconditions hold, \cref{cond:live} is violated for the declared model. \Cref{thm:sound} still guarantees safety, and completion then depends on diagnostic actions that the planner composes outside the declared propositions. We call missions with such a fault hidden-cause missions.

\begin{theorem}[Completion]
\label{thm:complete}
Let \cref{asm:evidence} and \cref{cond:obs,cond:live,cond:conn} hold, let $s_0\models K_0$, and let the planner be complete. Then:
\begin{enumerate}
\item[(i)] any run in which every actuation step executed under true preconditions realizes all effects with conclusive judgments terminates within $N=4|P|+2|\Asense|+N_x+2$ episodes,
\item[(ii)] every run terminates with probability one,
\end{enumerate}
ending either with $\kap_T\models G$ or in a hold certified by \cref{cor:hold}.
\end{theorem}

\begin{proof}
Fix any decision point and let $E$ denote the event that all subsequent actuation steps under true preconditions realize their full effects and yield conclusive judgments. On $E$, no judgment is $\evU$, so decided propositions outside $P_x$ remain decided. Define the potential $\Phi=|(P\setminus P_x)\cap\kap^{-1}(\U)|+|((P\setminus P_x)\cap\kap^{-1}(\U))\setminus\Fs|\le 2|P|$, which is non-increasing on $E$. Call an episode productive if $\Phi$ strictly decreases within it.

Consider an episode on $E$ free of exogenous changes and deadline expirations. Completeness and \cref{cond:conn} guarantee a valid plan $\pi$. While $\Phi$ remains unchanged, execution tracks the optimistic relaxation: sensing steps have verified preconditions and open deadlines, either deciding a proposition ($\Phi$ decreases) or observing decided ones; known-precondition actuation steps succeed by \cref{thm:sound}(i); and any reversible step with undecided preconditions $U^{-}$ passes line~\ref{ln:revsense} because $\pi$ assumes nothing in $\Fs$. Because telemetry has just decided $P_x$, $U^{-}$ is a non-empty subset of $(P\setminus P_x)\setminus\Fs$. If $U^{-}$ holds in truth, the step succeeds; if not, the physical state is unchanged, and either all effects already held or at least one is conclusively refuted, causing every element of $U^{-}$ to become decided or enter $\Fs$ and decreasing $\Phi$ by at least $|U^{-}|\ge1$. An unverified irreversible step either returns a certified hold (line~\ref{ln:hold}) or sets a pending cover $C$ (line~\ref{ln:irrsense}). If $\pi$ completes in agreement with the relaxation so that $\kap\models g$, then $g=G$ yields termination at the next telemetry check, whereas a route goal $g=\pre(\sigma)$ executes the head of $C$ at line~\ref{ln:pending} prior to the next planner call, deciding a proposition outside $P_x$ and decreasing $\Phi$.

Episodes on $E$ thus partition into at most $2|P|$ productive episodes, at most $|\Asense|$ deadline-expiration episodes, at most $N_x$ exogenous-change episodes, at most $1$ terminal episode, and non-productive episodes that set $C$ when $g=G$. Because $C$ remains non-empty across exogenous interruptions until an element executes (productive), a deadline discards $C$, or the run terminates, at most $2|P|+|\Asense|+1$ episodes set $C$, yielding $N=4|P|+2|\Asense|+N_x+2$ and proving (i).

For (ii), at most $L$ actuation steps execute per episode, so from any history $E$ holds across the next $N$ episodes with probability at least $(\rho q)^{LN}>0$; hence non-termination after $kN$ episodes has probability at most $(1-(\rho q)^{LN})^k\to0$.
\end{proof}

\Cref{thm:complete} accommodates non-monotonic knowledge loss from $\evU$ judgments through the one-attempt failure set $\Fs$ and measurable hardware rates $(\rho,q)$. Because the completeness of a foundation model cannot be assumed, the implementation imposes a skill budget and a planner-call cap, which leave \cref{thm:sound} intact.

% ======================================================================
\section{Experimental design}
\label{sec:design}

Two testbeds evaluate the three theorems and their assumptions. Both use Thoth, the executor that implements \AKR{}. Full-stack runs pair Thoth with a foundation-model planner in a physics-based simulation of a lunar rover. Mission-level Monte Carlo trials pair the same executor with a search planner in a sampled world.

\subsection{Simulation platform}
\label{sec:platform}

\begin{figure*}[t]
\centering
\includegraphics[width=\textwidth]{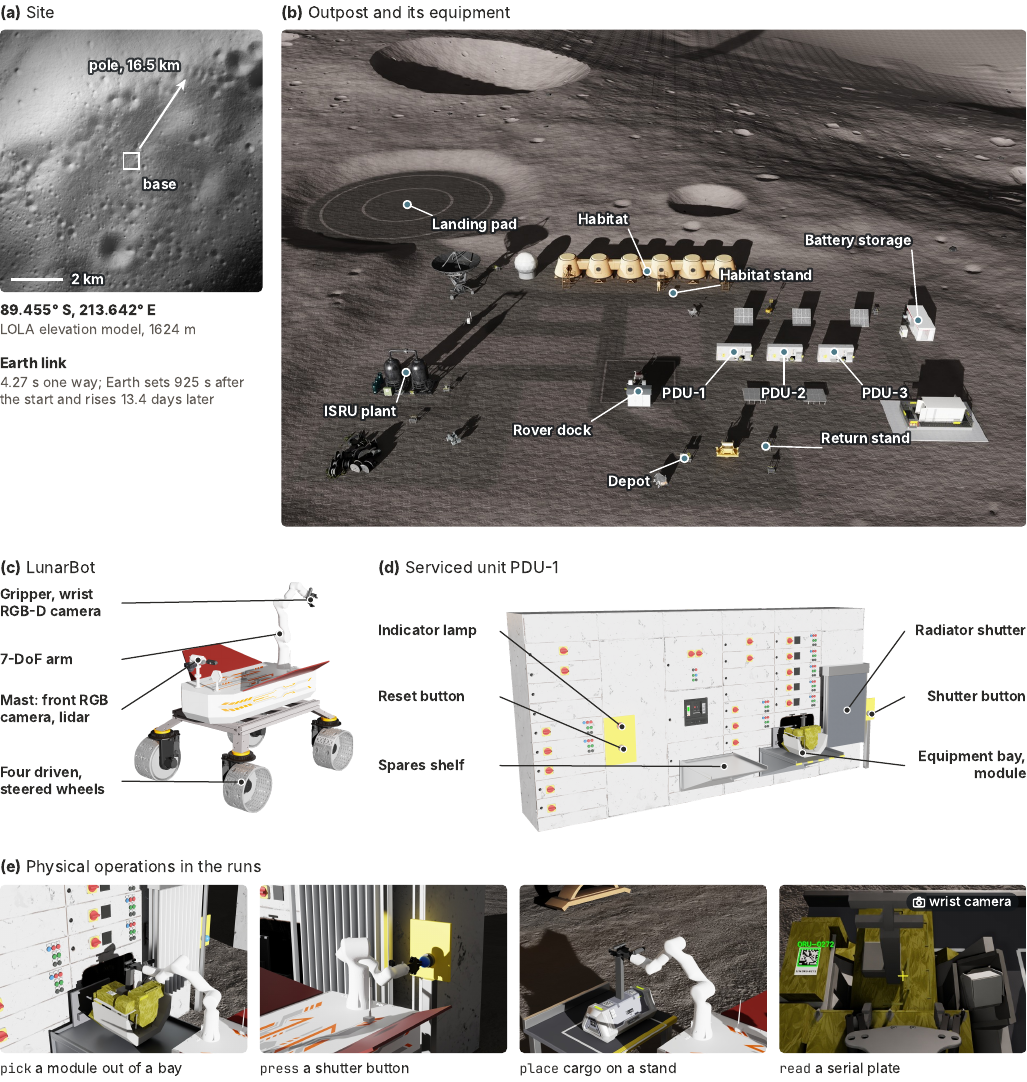}
\caption{Simulation platform: (a) the south-polar site, (b) the outpost and its equipment, (c) the rover LunarBot, (d) the interfaces of a serviced unit, shown on PDU-1, and (e) four physical operations from the full-stack runs.}
\label{fig:platform}
\end{figure*}

The surface outpost is implemented in Isaac Sim via Space Robotics Bench~\cite{orsula2025srb} using Lunar Orbiter Laser Altimeter terrain data (5~m/pixel) for the south-polar Connecting Ridge site ($89.455^{\circ}$\,S, $213.64^{\circ}$\,E, local slope $\approx2.5^{\circ}$)~\cite{barker2021lola} (\cref{fig:platform}(a)). The facility has three power distribution units (PDU-1, PDU-2, PDU-3), a battery energy storage system (BESS), an ISRU plant, thirteen supports, six equipment bays and a rover dock (\cref{fig:platform}(b)). It holds thirteen Data Matrix-labelled items: six ORUs including one permanently failed unit, two battery modules, two filter cartridges, a cargo crate, a consumables container and a return container. Contact interactions follow rigid-body dynamics, and equipment anomalies obey state-transition dynamics: resetting an overheated unit with its radiator or condenser shutter closed causes a thermal re-trip after 30~s, while resetting a unit containing an unserviceable module causes a re-trip after 3~s (\cref{fig:platform}(d)). The rover, LunarBot, combines a holonomic four-wheel-steered base, a seven-degree-of-freedom manipulator, a parallel gripper, lidar, a front RGB camera and a $640\times480$ wrist RGB-D camera over ROS~2~\cite{macenski2022ros2}, Nav2 map localization and MoveIt RGB-D pose servoing~\cite{dong2015pbvs} (\cref{fig:platform}(c) and (e)). Solar elevation is fixed at $12^{\circ}$. Gravity is lunar, 1.62\,m/s$^2$.

\subsection{Physical grounding of actions, sensing and evidence}
\label{sec:grounding}

Propositions $P$ encode equipment and object states ($|P|=355$ in power missions, $368$ in deliveries). The exogenous set $P_x$ contains each unit's breaker, power, indicator-lamp and over-temperature state and each bay's installed module, serial number and connector status, all read by telemetry. Navigation, docking, button presses, picks and placements at supports, arm postures, and module swaps at bays of de-energized units belong to $\Arev$. Two operations form $\Airr$. Removing a module from a bay of an energized unit requires the grant $\mathrm{granted\_remove}$ and permanently damages the connector. Because the power state of every unit belongs to $P_x$ and telemetry reads it before every action, a pick at a bay is known to be a swap or a removal before it is dispatched. Handing over cargo at its destination requires verified cargo identity and an unoccupied destination. Fault causes and the serviceability notes of the inventory record reach the planner as text outside $\kap$. No declared proposition links a closed shutter or an unserviceable module to a re-trip, so the five contingency repairs and the sequential-fault repair are the hidden-cause missions of \cref{sec:prop-complete}.

\Cref{tab:sensing} lists the sensing actions; their costs are skill medians measured before the campaign. Camera costs count frame acquisition ($0.02$~s), because the mission clock is held during inference (\cref{sec:executor}). A serial-plate read requires the rover docked at the support with empty jaws and decides both occupancy and item identity. The held-item camera detects whether the gripper holds something but not which item, so identifying a held item requires a held-item check: placing it on a free support, reading its plate and picking it up again.

\begin{table}[t]
\caption{Sensing actions and their physical realization.}
\label{tab:sensing}
\centering
\footnotesize
\setlength{\tabcolsep}{3pt}
\begin{tabularx}{\columnwidth}{@{}>{\raggedright\arraybackslash}p{2.15cm}>{\raggedright\arraybackslash}Xcc@{}}
\toprule
Sensing action & Realization & Cost (s) & Deadline (s) \\
\midrule
Telemetry & equipment state service & 0 & none \\
Localization, arm state & pose estimate, joint encoders & 0 & none \\
Camera question & RGB image, vision-language call & 0.02 & none \\
Held-item camera & wrist camera (holding or empty) & 0.02 & none \\
Serial-plate read & arm viewpoint, Data Matrix decode & 5.7 & none \\
Held-item check & put down, read plate, pick up & 153.6 & none \\
Ground request & console request over Earth link & 308.5 & $T_{\mathrm{vis}}-308.5$ \\
\bottomrule
\end{tabularx}
\end{table}

Skill-specific evidence rules (\cref{tab:evidence}) compute $\nu_a(e)$ from physical measurements only and never use a controller's completion flag. Navigation and docking evaluate pose estimates against $0.25$~m/$0.30$~rad and $0.06$~m/$0.06$~rad bands; picking evaluates jaw closure, $\ge0.05$~m lift and optical flow; placement and handover evaluate jaw release, $\ge0.10$~m withdrawal and settled support visibility. Button pressing verifies contact travel from arm kinematics and assigns $\evU$ to the equipment effect, which telemetry or a camera question then decides.

\begin{table}[t]
\caption{Evidence rules and the measurements that decide them.}
\label{tab:evidence}
\centering
\footnotesize
\setlength{\tabcolsep}{3pt}
\begin{tabularx}{\columnwidth}{@{}l>{\raggedright\arraybackslash}X>{\raggedright\arraybackslash}X@{}}
\toprule
Action & Measurements & Unverified when \\
\midrule
Navigate, dock & localization pose against band & pose estimate absent \\
Press & button contact depth, retreat & equipment effect (decided by telemetry or camera) \\
Pick & jaw angle, lift, image flow & flow inconclusive or item unknown \\
Place, hand over & jaw open, withdrawal, settled view & destination out of view or item unknown \\
Stow, arm ready & joint encoders & never \\
\bottomrule
\end{tabularx}
\end{table}

\subsection{Earth-link geometry and ground authorization}
\label{sec:link}

Direct-to-Earth link geometry was computed via SPICE~\cite{acton1996spice} using the JPL DE440 ephemeris for 11 October 2028. Combining $1.271$~s one-way light time with $3$~s relay and ground-processing overhead per leg yields a $4.27$~s one-way delay; adding $300$~s for flight-controller review gives
\begin{equation}
c_g = 2\times 4.27~\mathrm{s} + 300~\mathrm{s} = 308.54~\mathrm{s}\approx 308.5~\mathrm{s}.
\label{eq:cg}
\end{equation}
Lunar libration carries Earth below the Connecting Ridge horizon on 11 October 2028 for $13.4$~days~\cite{mazarico2011illumination}. Fifteen missions set $T_{\mathrm{vis}}=925$~s ($w_{\sigma_a}\approx616.5$~s). In the no-spare repair, base and inserted-proposal runs start earlier in the pass at $T_{\mathrm{vis}}=1543$~s ($w_{\sigma_a}\approx1234.5$~s), whereas an early-Earth-set condition sets $T_{\mathrm{vis}}=925$~s ($w_{\sigma_a}\approx616.5$~s). On the rover, a ground request proposed by the planner runs in the background while the rover moves. If a removal is proposed while the request is in flight, Thoth withholds it until the reply arrives.

\subsection{Planner and executor implementation}
\label{sec:executor}

The planner in the full-stack runs is the foundation model DeepSeek-V4.1-Flash with reasoning enabled. Each call is stateless: its context is rebuilt from the event log and contains the mission brief, the goal, $\kap$, the remaining record entries $R_{\mathrm{cur}}$, the executor's feedback and the link status. The model receives typed skill interfaces and no mapping from faults to remedies. It maintains an explicit working plan and returns one tool call per call. Consecutive steps under an unchanged working plan form one episode of \cref{alg:akr}, with $L=16$. A run ends at a budget of $80$ skills or a cap of $200$ planner calls, and it fails when the planner reports that the mission is impossible. The planner is assumed to run on computing resources local to the outpost, so no planner call depends on the Earth link, and the hosted model used here stands in for such a resource. The mission clock is held during the $6994$ planner calls and the $1112$ vision calls of the campaign, which the same model answers, so that mission times do not depend on the computing hardware. Mission time therefore counts rover execution and the $308.5$~s ground request and excludes inference latency. In the median, a planner call took 16.1~s, a camera question 18.5~s, and all planner calls of one run 744~s under \AKR{}, 1080~s under the \KR{} and 681~s under the \FFR{}.

\subsection{Mission suite and compared methods}
\label{sec:missions}

\Cref{tab:missions} defines the sixteen maintenance and delivery missions in four classes: seven reference missions, six contingency missions, one sequential-fault mission and two wrong-record missions. The nine missions outside the reference class are the non-reference missions. An external evaluator judges success from simulator truth with the same predicate for every method. A repair succeeds when the faulted unit is restored with no damaged connector. A delivery succeeds when the cargo rests on its destination and, where the destination was occupied, the occupier rests on another support. Both require the rover parked at the dock with the arm stowed.

\begin{table*}[t]
\caption{Goals and conditions of the sixteen missions in four classes. A dagger marks the six hidden-cause missions.}
\label{tab:missions}
\centering
\footnotesize
\setlength{\tabcolsep}{6pt}
\renewcommand{\arraystretch}{1.08}
\begin{tabularx}{\textwidth}{@{}>{\raggedright\arraybackslash}p{0.33\textwidth}>{\raggedright\arraybackslash}X@{}}
\toprule
Goal & Condition \\
\midrule
\multicolumn{2}{@{}l}{\textit{Reference missions (7)}} \\
Restore PDU-2 & Overcurrent breaker trip at PDU-2; a reset restores the unit. \\
Restore BESS & Overcurrent breaker trip at BESS; a reset restores the unit. \\
Restore ISRU & Overcurrent breaker trip at ISRU; a reset restores the unit. \\
Locate and restore faulty equipment & Brief reports PDU-2; BESS is the unit tripped on overcurrent. \\
Deliver spare ORU to PDU-3 & ORU-0417 from depot slot~0 to the PDU-3 spares shelf. \\
Deliver battery module to BESS & BAT-0102 from the depot spares stand to the BESS receiving stand. \\
Deliver consumables to habitat & CNS-0301 from landing slot~0 to the habitat logistics stand. \\
\midrule
\multicolumn{2}{@{}l}{\textit{Contingency missions (6)}} \\
Restore PDU-1 (shutter closed)$^{\dagger}$ & Overheating trip ($85\,^{\circ}\mathrm{C}$) with radiator shutter closed; reset alone re-trips after 30~s. \\
Restore BESS (shutter closed)$^{\dagger}$ & Overheating trip ($85\,^{\circ}\mathrm{C}$) with condenser shutter closed; reset alone re-trips after 30~s. \\
Restore BESS (failed module)$^{\dagger}$ & Installed battery BAT-0101 unserviceable; reset re-trips after 3~s until spare BAT-0102 is installed. \\
Restore ISRU (clogged filter)$^{\dagger}$ & Installed filter FLT-0201 clogged; spare FLT-0202 must be installed before reset holds. \\
Restore PDU-1 (no spare)$^{\dagger}$ & ORU-0211 failed and sole spare ORU-0417 unserviceable; compatible modules sit in energized units and require ground authorization to remove. \\
Deliver consumables (destination occupied) & Habitat stand occupied by return container RTN-0007, which must be relocated to another support first. \\
\midrule
\multicolumn{2}{@{}l}{\textit{Sequential-fault mission (1) and wrong-record missions (2)}} \\
Restore BESS (sequential fault)$^{\dagger}$ & Failed battery module followed by an overheating trip after replacement and reset because the shutter is closed. \\
Deliver serial item (mislocated) & ORU-0417 to PDU-3 shelf; recorded at depot slot~0 (which holds failed ORU-0272) but physically on PDU-1 shelf. \\
Deliver serial item (records exchanged) & ORU-0417 to PDU-1 shelf; locations of ORU-0417, ORU-0272 and CRT-0112 permuted across three supports. \\
\bottomrule
\end{tabularx}
\end{table*}

Three acting rules and two baselines are compared. The acting rules run inside Thoth with the same foundation-model planner, skills and evidence rules, and they differ only in admissibility (\cref{tab:strategies}). The \KR{} requires verified preconditions before every action and sets a cover when one is undecided~\cite{bonet2011kreplanner,brafman2012sdr}. The \FFR{} admits any action without a known-false precondition~\cite{yoon2007ffreplan,yao2023react}. \AKR{} applies the first condition to irreversible actions and the second to reversible actions, where each proposition is attempted unverified at most once. The two baselines represent flown space autonomy. PLEXIL~6.1.3~\cite{estlin2006plexil} executes hand-written procedures, and \EP{}~\cite{barreiro2012europa} plans in an authored domain model and dispatches through PLEXIL. Both repeat a failed call at most three times, then hold and request a new procedure from the ground. For deliveries that name a serial number, their procedures include a serial-plate search written before the campaign. Each baseline ran in its original form on all sixteen missions and in a revised form on the six contingency missions. The revised forms were written after each contingency class had been diagnosed, with at most 3~h of active engineering time per class.

\begin{table}[t]
\caption{Admissibility conditions of the three acting rules.}
\label{tab:strategies}
\centering
\footnotesize
\setlength{\tabcolsep}{3pt}
\begin{tabularx}{\columnwidth}{@{}l>{\raggedright\arraybackslash}X>{\raggedright\arraybackslash}X@{}}
\toprule
Acting rule & Reversible action & Irreversible action \\
\midrule
\KR{} & known true, after cover & known true, after cover \\
\FFR{} & not known false & not known false \\
\AKR{} & not known false, each guess once & known true, after cover \\
\bottomrule
\end{tabularx}
\end{table}

\subsection{Experimental protocol and metrics}
\label{sec:metrics}

The evaluation consists of 165 full-stack runs (\cref{tab:campaign}) and 254,000 Monte Carlo trials. All full-stack runs used one frozen prompt. Runs interrupted by a simulator or supervision fault were repeated in the same configuration. Every run in which the planner gave up, perception erred or a manipulation failed is kept with its outcome.

\begin{table}[t]
\caption{Full-stack runs: 121 runs of the three acting rules and 44 baseline runs.}
\label{tab:campaign}
\centering
\footnotesize
\setlength{\tabcolsep}{3pt}
\begin{tabularx}{\columnwidth}{@{}Ylc@{}}
\toprule
Configuration & Missions & Runs \\
\midrule
\AKR{}, base condition & all 16 & 48 \\
\KR{}, base condition & 9 non-reference & 27 \\
\FFR{}, base condition & 9 non-reference & 27 \\
\AKR{} (7) and \FFR{} (9), inserted proposals & 2 & 16 \\
\AKR{}, early Earth set ($T_{\mathrm{vis}}=925$~s) & 1 (no-spare) & 3 \\
PLEXIL and \EP{}, original & all 16 & 32 \\
PLEXIL and \EP{}, revised & 6 contingency & 12 \\
\bottomrule
\end{tabularx}
\end{table}

\Cref{sec:results} uses four quantities. Evaluator success is defined in \cref{sec:missions}. A hazardous action is an irreversible action dispatched to the robot with at least one precondition false in simulator truth. It is counted once per execution, and a withheld proposal is not counted. A decision point is a proposal of an irreversible action with at least one undecided precondition; there, the sensing cost paid is compared with the lower bound $\cst_D(U,t)$ of \cref{thm:cost}. The time ratio of a run is its mission time $t_j$ divided by the shortest time $t^{\min}_{m(j)}$ of any successful run on the same mission under any acting rule, with $t_j=\infty$ for an unsuccessful run. Its distribution over the $n_s$ runs $J_s$ of rule $s$ is the performance profile of Dolan and Mor\'e~\cite{dolan2002profiles},
\begin{equation}
\phi_s(\tau)=\frac{1}{n_s}\,\bigl|\{\,j\in J_s:\ t_j\le\tau\,t^{\min}_{m(j)}\,\}\bigr| .
\label{eq:profile}
\end{equation}

\paragraph{Full-stack protocol}
In the base condition each acting rule runs every mission three times: \AKR{} on all sixteen missions, the \KR{} and the \FFR{} on the nine non-reference missions. Two further conditions test \cref{thm:sound} and \cref{cor:hold}. The inserted-proposal condition, run under \AKR{} and the \FFR{}, places an irreversible action at the front of the planner's sequence while one of its preconditions is undecided. In the no-spare repair the inserted action is a removal at an energized bay, and the ground is set to refuse the grant. In the destination-occupied delivery it is the handover, inserted once the rover is at the destination with the cargo held; runs were launched until the insertion had fired in three runs per rule. The early-Earth-set condition runs the no-spare repair under \AKR{} with $T_{\mathrm{vis}}=925$~s.

\paragraph{Monte Carlo trials}
The Monte Carlo trials keep Thoth unchanged and replace the rover and the simulator by a sampled world that follows the same equipment rules. An actuation action under true preconditions realizes all its effects with probability $\rho=0.847$, and each of its effect judgments is conclusive with probability 0.962, independently of the others, so an action with $k$ effects meets \cref{cond:live} with $q=0.962^{k}$. Both rates are pooled over the actions on the rover (\cref{sec:res-evidence}). Each sensed value is flipped with probability $\eta$, and each record entry is false with probability $p_R$.

The planner is a breadth-first search planner that returns a shortest valid plan, and it stands in for the complete planner of \cref{thm:complete}. It plans in the optimistic relaxation under \AKR{} and the \FFR{} and without assumptions on actuation preconditions under the \KR{}. With probability $\varepsilon$, a planner call whose plan contains an irreversible action is faulted: the irreversible action is moved to the front, or the sensing before it is deleted. Ten missions are solvable from the declared propositions. In the six hidden-cause missions the search planner is told the remedy and $\kap$ is not, so these trials test the executor and not diagnosis.

Three grids are run. The main grid tests \cref{thm:sound,thm:cost}. It sweeps $\varepsilon$ from 0 to 1 at $p_R=0.1$ and $\eta=0$, $p_R$ from 0 to 1 at $\varepsilon=\eta=0$, and $\eta$ from 0.001 to 0.1 at $\varepsilon=p_R=0$, which gives 28 points with 1000 trials per point and rule (84,000 trials). The ground-refusal grid tests \cref{thm:sound}(iii) and \cref{cor:hold}. On the six hidden-cause missions the ground refuses each request with probability $0.5$ at the same 28 points (168,000 trials, with 28,000 no-spare trials per rule). The episode grid tests \cref{thm:complete}. \AKR{} runs the ten solvable missions without injected errors at $\rho=0.847$ and at $\rho=0.4235$, with 1000 trials each.

% ======================================================================
\section{Results}
\label{sec:results}

The three theorems make three measurable predictions. \Cref{thm:sound} predicts no hazardous action under sound observations at any planner fault rate and any record error rate. \Cref{thm:cost} predicts that the sensing paid at each decision point equals the minimum, and that it meets the lower bound unless an actuation effect can replace a sensor. \Cref{thm:complete} predicts that, with a complete planner, every run ends with the goal verified or in a certified hold, within $N$ episodes when no action fails. Each prediction is tested first in the Monte Carlo trials, where the conditions of its theorem hold, and then on the full stack. \Cref{sec:res-evidence} measures how far the rover departs from \cref{asm:evidence} and \cref{cond:live}.

\subsection{Safe execution under wrong records and faulty proposals}
\label{sec:res-sound}

\begin{figure*}[t]
\centering
\includegraphics[width=\textwidth]{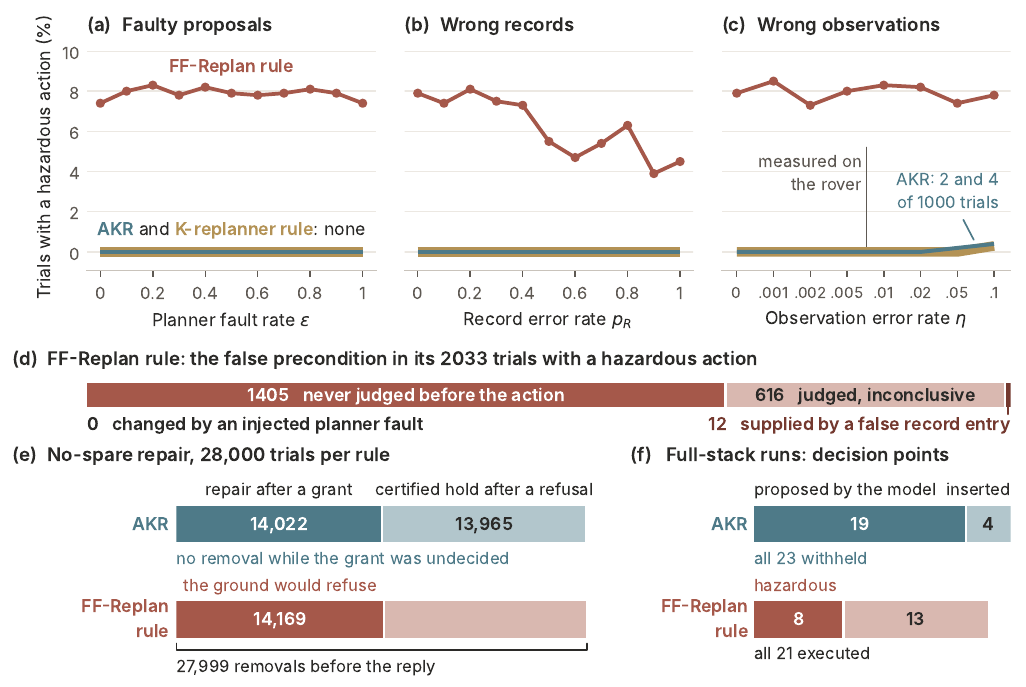}
\caption{Safe execution under the three acting rules. (a) to (c) Trials with a hazardous action in the main grid, with 1000 trials per point and rule, against the planner fault rate $\varepsilon$, the record error rate $p_R$ and the observation error rate $\eta$; the tested values of $\eta$ are spaced equally, and the vertical line marks the rate measured on the rover. (d) What was known of the false precondition in the 2033 trials of the \FFR{} with a hazardous action. (e) The no-spare trials of the ground-refusal grid. (f) Decision points in the full-stack runs.}
\label{fig:safety}
\end{figure*}

\Cref{thm:sound}(ii) holds for any planner and any record, and the main grid takes both quantifiers to their extremes. At $\varepsilon=1$ every plan that contains an irreversible action is faulted, and at $p_R=1$ every record entry is false. \AKR{} executed no hazardous action in the 21,000 trials with $\eta=0$ (\cref{fig:safety}(a) and (b)). The \KR{} also executed none, and its rule for irreversible actions is the irreversible branch of \eqref{eq:adm}. Safety therefore comes from that branch alone, and the optimistic reversible branch of \AKR{} costs none of it.

The \FFR{} shows what the irreversible branch prevents. It executed a hazardous action in 7.4\% to 8.3\% of trials at every planner fault rate, and in 7.9\% of trials at $p_R=0$ against 4.5\% at $p_R=1$. All 2033 of its trials with a hazardous action in the main grid were replayed. In 1405 of them the false precondition had never been judged, in 616 an inconclusive judgment had left it undecided, and in 12 a false record entry supplied it. No false precondition came from an injected planner fault (\cref{fig:safety}(d)). Of the 333 hazardous actions in faulted planner calls, 325 came from plans that lacked sensing before the irreversible action even without the fault. The cause lies in the plans made under this rule. At $\varepsilon=0$ only 48 of its 1938 plans with an irreversible action place sensing before it. A rule that asks for ``not known false'' admits an undecided precondition, and the irreversible branch of \eqref{eq:adm} asks for ``known true'' and withholds exactly this class. The 616 trials tie the result to the three-valued evidence of \cref{sec:variant}: an inconclusive judgment leaves the proposition undecided, and \AKR{} then demands a sensing action before it acts.

\Cref{thm:sound}(iii) states that a restricted action executes only after its grant has been observed. In the 28,000 no-spare trials of the ground-refusal grid, \AKR{} never removed the module while the grant was undecided (\cref{fig:safety}(e)). The ground refuses with probability $0.5$, and the outcomes split accordingly: 14,022 trials completed the repair after a grant, and 13,965 ended in a certified hold after a refusal. The \FFR{} removed the module before the reply arrived in 27,999 of its 28,000 trials, and in 14,169 of them the ground would have refused. \AKR{} needs no separate rule for ground authorization. The grant is a precondition that only a ground request can observe (\cref{sec:variant}), so the admissibility check that withholds a handover onto an unread support also withholds a removal without a grant.

The full-stack runs show that the quantifier ``any planner'' is needed in practice. In the 58 \AKR{} runs the foundation model proposed an irreversible action with an undecided precondition 19 times while working normally, and the protocol inserted four more. Thoth withheld all 23, and no \AKR{} run contains a hazardous action (\cref{fig:safety}(f), \cref{tab:safety}). The \FFR{} executed all 21 of its decision points, and 8 of these actions were hazardous: six handovers onto the occupied habitat stand and two removals at the energized PDU-3 bay. After each of the six handovers the cargo came to rest on the ground, where no skill can pick it up. The irreversible branch therefore also protected completion. On the destination-occupied delivery, \AKR{} completed 5 of 7 runs and the \FFR{} 2 of 9. A withheld handover cost one serial-plate read of 5.7~s in 17 of 19 cases.

\begin{table}[t]
\caption{Decision points and hazardous actions in the full-stack runs. The upper block pools the no-spare repair and the three non-reference deliveries; the lower block counts all runs of the two rules.}
\label{tab:safety}
\centering
\footnotesize
\setlength{\tabcolsep}{3pt}
\begin{tabularx}{\columnwidth}{@{}Ycccc@{}}
\toprule
 & \multicolumn{2}{c}{Base condition} & \multicolumn{2}{c}{Inserted proposals} \\
\cmidrule(lr){2-3}\cmidrule(l){4-5}
 & \AKR{} & \makecell{FF-Replan\\rule} & \AKR{} & \makecell{FF-Replan\\rule} \\
\midrule
Runs & 12 & 12 & 7 & 9 \\
Decision points withheld & 8 & 0 & 4 & 0 \\
Covers set & 7 & 0 & 3 & 0 \\
Median cover cost (s) & 5.7 & n/a & 5.7 & n/a \\
Removals without an observed grant & 0 & 2 & 0 & 2 \\
Hazardous actions & 0 & 2 & 0 & 6 \\
\subrow removal at an energized bay & 0 & 0 & 0 & 2 \\
\subrow handover onto the occupied destination & 0 & 2 & 0 & 4 \\
Certified holds & 0 & 0 & 3 & 3$^{\mathrm{a}}$ \\
Evaluator successes & 10 & 9 & 3 & 1 \\
\midrule
 & \multicolumn{2}{c}{\AKR{}} & \multicolumn{2}{c}{FF-Replan rule} \\
\cmidrule(lr){2-3}\cmidrule(l){4-5}
All runs & \multicolumn{2}{c}{58} & \multicolumn{2}{c}{36} \\
Decision points & \multicolumn{2}{c}{23} & \multicolumn{2}{c}{21} \\
\subrow withheld & \multicolumn{2}{c}{23} & \multicolumn{2}{c}{0} \\
\subrow executed & \multicolumn{2}{c}{0} & \multicolumn{2}{c}{21} \\
Hazardous actions & \multicolumn{2}{c}{0} & \multicolumn{2}{c}{8} \\
\bottomrule
\multicolumn{5}{@{}l}{\footnotesize $^{\mathrm{a}}$Two of these runs had already executed a hazardous removal.}
\end{tabularx}
\end{table}

\Cref{thm:sound} assumes sound observations, and the $\eta$ sweep measures how far that assumption can fail before the conclusion does. \AKR{} executed no hazardous action in the 5000 trials with $\eta$ from 0.001 to 0.02, and the last value is nearly three times the rate of 0.0072 measured on the rover (\cref{fig:safety}(c)). At $\eta=0.05$, 2 of 1000 trials contained a hazardous action, and at $\eta=0.1$, 4 of 1000. The structure of the theorem explains the margin: a wrong observation leads to a hazardous action only when it decides a precondition of an irreversible action. Completion is unaffected up to $\eta=0.01$, where \AKR{} completed all 627 trials on the solvable missions at every point.

\subsection{The cost of safety and sensing on demand}
\label{sec:res-sensing}

\begin{figure*}[t]
\centering
\includegraphics[width=\textwidth]{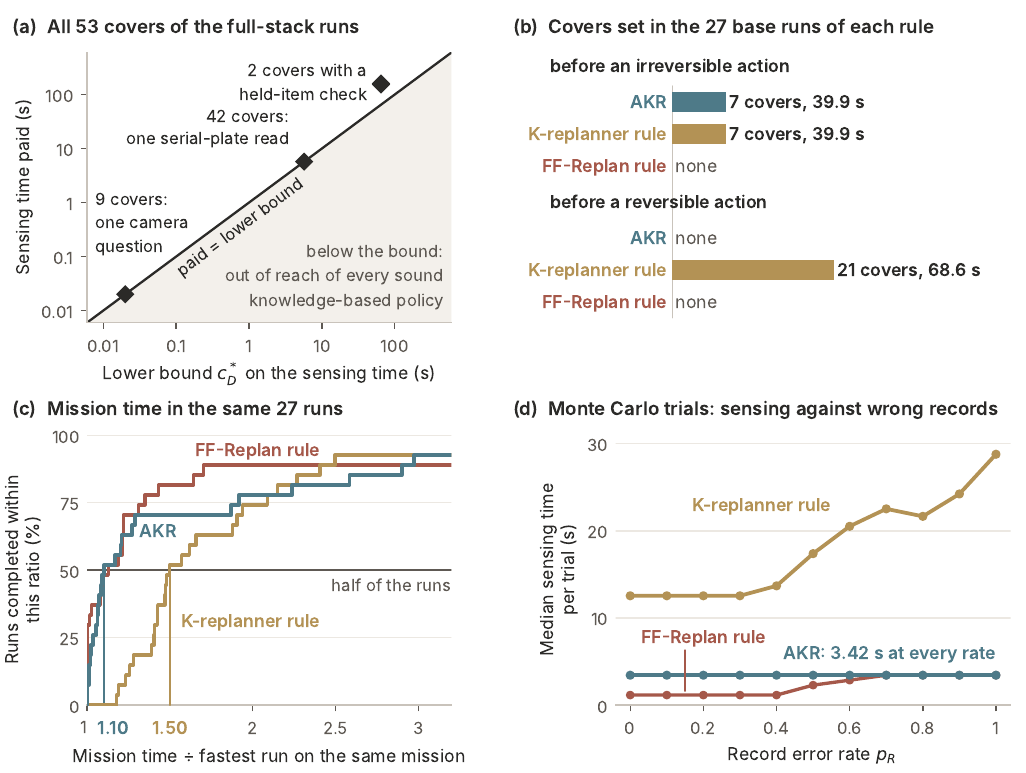}
\caption{Cost of the three acting rules. (a) Sensing time paid against the lower bound $\cst_D$ of \cref{thm:cost} for the 53 covers of the full-stack runs, on logarithmic axes. (b) Covers set in the 27 base runs of each rule on the nine non-reference missions, by the class of the action that demanded them, with the sensing time paid. (c) Performance profiles of mission time in the same runs; the vertical lines mark the median time ratios of \AKR{} and the \KR{}. (d) Median sensing time per trial on the ten solvable missions of the main grid against the record error rate $p_R$.}
\label{fig:sensing}
\end{figure*}

\Cref{thm:cost}(i) and (ii) bound the sensing before an irreversible action from below for every sound knowledge-based policy. Under \AKR{} and the \KR{}, Thoth set 53 covers in the full-stack runs, and each paid the exact minimum $\cst(U,t)$. \AKR{} set 19 of them before an irreversible action, and 17 of these paid the lower bound $\cst_D(U,t)$ with a single serial-plate read of 5.7~s. At these 17 decision points no sound knowledge-based policy could have paid less, so this sensing is the price of safety and not an overhead of the algorithm. The two exceptions are handovers whose cover needed a held-item check, and they paid 159.3~s and 153.6~s against a bound of 64.7~s (\cref{fig:sensing}(a)). They are the case that \cref{thm:cost}(ii) excludes from equality: the bound admits an actuation action as a sensor at its duration, and the executor covers with sensing actions only.

The asymmetry of \eqref{eq:adm} removes only sensing that \cref{thm:cost} does not require. In the 27 base runs on the non-reference missions, \AKR{} and the \KR{} each set 7 covers before irreversible actions and paid 39.9~s for them (\cref{tab:price}, \cref{fig:sensing}(b)). The \KR{} set 21 further covers before reversible actions, and \AKR{} set none. The two rules thus pay the same for safety and differ only before reversible actions, where a wrong assumption costs time and no hardware. The median time ratio is 1.10 under \AKR{} and 1.50 under the \KR{} (\cref{fig:sensing}(c)). The 21 reversible covers cost 68.6~s of sensing in total, which is too little to account for this difference. The difference appears in the actions around them: the median \KR{} run executed 32 actuation actions and the median \AKR{} run 24.

\begin{table}[t]
\caption{The three acting rules on the nine non-reference missions (27 base runs per rule).}
\label{tab:price}
\centering
\footnotesize
\setlength{\tabcolsep}{3pt}
\begin{tabularx}{\columnwidth}{@{}Yccc@{}}
\toprule
 & \AKR{} & \makecell{K-replanner\\rule} & \makecell{FF-Replan\\rule} \\
\midrule
Evaluator successes (of 27) & 25 & 25 & 24 \\
\midrule
\multicolumn{4}{@{}l}{\textit{Medians over the 27 runs}} \\
\subrow Mission time (s) & 991.3 & 1097.5 & 800.9 \\
\subrow Time ratio & 1.10 & 1.50 & 1.13 \\
\subrow Actuation actions & 24 & 32 & 21 \\
\subrow Sensing actions & 3 & 5 & 3 \\
\subrow Sensing time (s) & 17.1 & 22.8 & 11.4 \\
\subrow Planner calls & 47 & 59 & 35 \\
\midrule
\multicolumn{4}{@{}l}{\textit{Covers, totals over the 27 runs}} \\
\subrow Before irreversible actions & 7 & 7 & 0 \\
\subrow Sensing time paid for them (s) & 39.9 & 39.9 & 0 \\
\subrow Before reversible actions & 0 & 21 & 0 \\
\subrow Sensing time paid for them (s) & 0 & 68.6 & 0 \\
\midrule
Irreversible covers at $\cst_D$, all runs & 17/19 & 7/7 & n/a \\
\bottomrule
\end{tabularx}
\end{table}

A wrong record costs \AKR{} an episode and no sensing. As $p_R$ rises from 0 to 1 in the main grid, the median sensing time of \AKR{}, averaged over the ten solvable missions, stays at 3.42~s (\cref{fig:sensing}(d)), and its median episode count rises from 4 to 8. A false entry is found by one failed reversible attempt, which ends the episode. The failure set then admits no second unverified attempt on the same proposition, so the number of such episodes is bounded. This is the role of the potential $\Phi$ in the proof of \cref{thm:complete}. The \KR{} must decide each unverified location by sensing before it acts, and its sensing time rises from 12.54~s to 28.79~s.

Uniform conservatism also removes plans. With $\eta=0$, 328 trials under the \KR{} ended at the planner-call cap, and in 319 of them the jaws state or the held item was undecided. A serial-plate read needs empty jaws, and a held-item check needs a support known to be free. The \KR{} requires known-true preconditions also for the reversible placement that would empty the jaws, so no sensing action is reachable: an exhaustive search found no valid plan in 29 of 30 sampled states. Its completions on the solvable missions fall from 626 of 627 trials at $\varepsilon=0$ to 561 at $\varepsilon=1$. \AKR{} may place onto a support of undecided occupancy, and none of its trials with $\eta=0$ ended at the planner-call cap. The solvability that \cref{cond:conn} demands thus depends on the optimistic reversible branch.

\Cref{thm:cost} bounds the sensing actions and not the time that the planner spends reaching them. The median mission time on the nine non-reference missions is 991.3~s under \AKR{} and 800.9~s under the \FFR{}. Five of these missions contain no decision point. On four of them the median \AKR{} run is shorter than the median \FFR{} run, and on the fifth it is 6\% longer. The largest difference is on the records-exchanged delivery, 1177.1~s against 412.0~s. Its cover is one read of 5.7~s, but the read needs empty jaws and the planner had already picked up the cargo. It put the item back, read the plate and picked the item up again, which took 483 to 495~s. In the no-spare repair \AKR{} waited for the ground's reply before the removal and the \FFR{} removed the module before the reply; the median successful run took 806.8~s and 693.5~s. The precondition of a cover element returns to the planner as a reversible subgoal (\cref{sec:cover}). The time to reach it depends on the planner's ordering of actions, which \eqref{eq:adm} does not constrain.

\begin{figure*}[t]
\centering
\includegraphics[width=\textwidth]{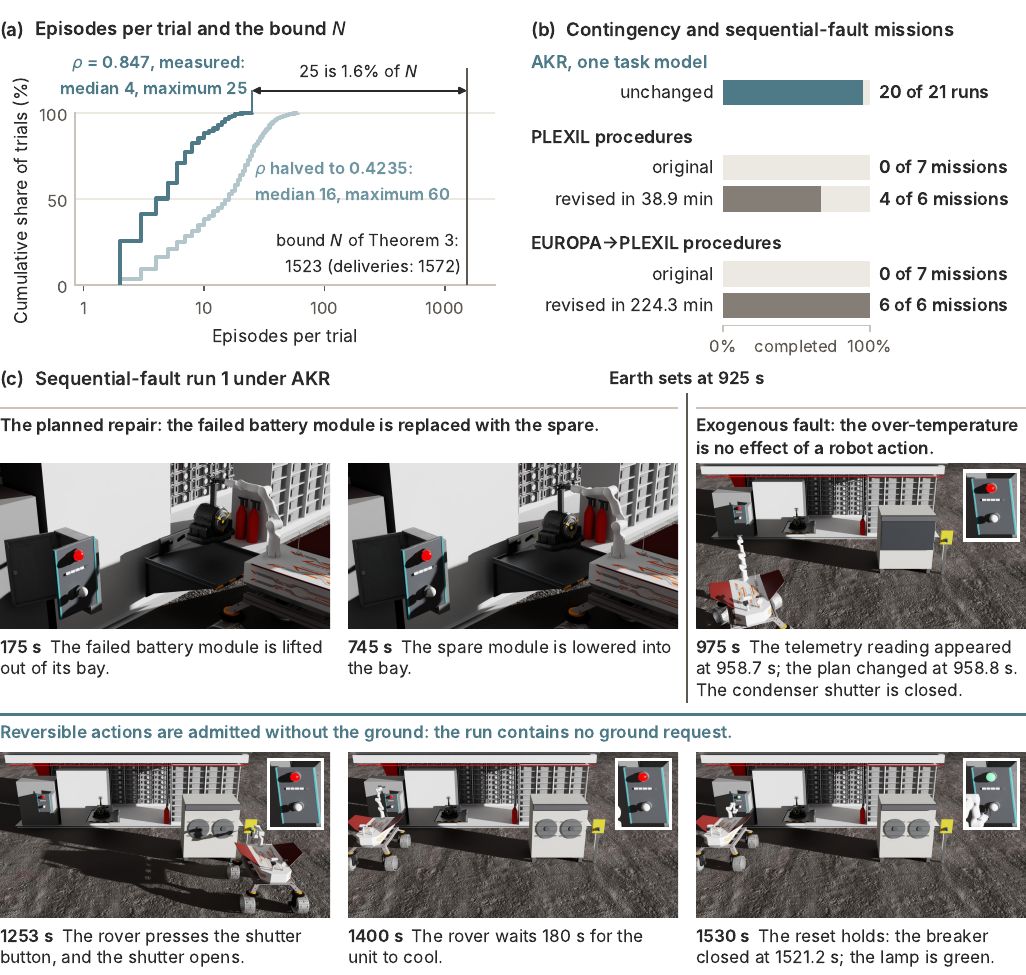}
\caption{Recovery and termination. (a) Cumulative distribution of the episodes per trial in the episode grid at $\rho=0.847$ and at $\rho=0.4235$, with the bound $N$ of \cref{thm:complete}. (b) Completion on the contingency and sequential-fault missions, counted in runs for \AKR{} and in missions for the original and the revised procedures. (c) Six frames replayed from sequential-fault run~1 under \AKR{}, each with its mission time; the insets enlarge the indicator lamp of the unit.}
\label{fig:timeline}
\end{figure*}

\subsection{Recovery and termination}
\label{sec:res-recovery}

\Cref{thm:complete}(i) bounds the episodes of a run without action failures by $N$, which is 1523 for power missions and 1572 for deliveries, and (ii) gives termination with probability one when actions fail. In the episode grid at the measured $\rho=0.847$, all 1000 trials ended with the goal verified, with a median of 4 episodes and a maximum of 25, which is 1.6\% of $N$ (\cref{fig:timeline}(a)). $N$ counts every proposition as if each had to be decided in an episode of its own, and a mission touches few of them. Action failures drive the episode count, as the proof implies: with $\rho$ halved to 0.4235 the median rose to 16 and the maximum to 60. This rate lies below the full-effect rate measured for every skill on the rover, the lowest being 0.517 for press. At that rate 991 trials completed and 9 ended at the budget of 80 skills. These 9 do not contradict \cref{thm:complete}(ii), which assumes no budget.

\Cref{thm:complete} assumes a complete planner, which cannot be proved for a foundation model. The hidden-cause missions sharpen the question: there the declared model violates \cref{cond:live} (\cref{sec:prop-complete}), and completion rests on the planner alone. The full-stack runs measure how often the foundation model filled this gap. \AKR{} completed all 18 base runs on the six hidden-cause missions and 20 of the 21 contingency and sequential-fault runs (\cref{fig:timeline}(b)). \Cref{tab:failures} lists how every run ended, and each ending without success corresponds to one premise of the theory. No \AKR{} run contains a hazardous action, because \cref{thm:sound} does not depend on the planner. In five runs an item came to rest off any support; drops occurred 10 times in 196 pick and place executions. No skill picks from the ground, so the state left the solvable set of \cref{cond:conn} and the planner gave up. Four runs ended in a certified hold with no irreversible action executed, as \cref{cor:hold} describes. In one run the task was complete in simulator truth but its evidence could not be obtained, and Thoth did not report success, because success requires $\kap\models G$.

\begin{table}[t]
\caption{How the 94 runs of \AKR{} and the \FFR{} ended.}
\label{tab:failures}
\centering
\footnotesize
\setlength{\tabcolsep}{3pt}
\begin{tabularx}{\columnwidth}{@{}Ycc@{}}
\toprule
Ending & \AKR{} & \makecell{FF-Replan\\rule} \\
\midrule
Runs & 58 & 36 \\
Success, goal verified & 48 & 25 \\
Certified hold after a refusal & 3 & 3$^{\mathrm{a}}$ \\
Certified hold on a late request & 1 & 0 \\
\midrule
\multicolumn{3}{@{}l}{\textit{Planner gave up}} \\
\subrow item at rest off any support & 5 & 1 \\
\subrow after a hazardous handover onto the occupied destination & 0 & 6 \\
\subrow goal still reachable & 0 & 1 \\
\subrow task complete, evidence unobtainable & 1 & 0 \\
\bottomrule
\multicolumn{3}{@{}p{\columnwidth}@{}}{\footnotesize $^{\mathrm{a}}$Two of these runs had already executed a hazardous removal.}
\end{tabularx}
\end{table}

\Cref{fig:timeline}(c) shows one sequential-fault run and the two parts of the model that carried it. The first is the exogenous set $P_x$. The over-temperature after the module replacement is no effect of a robot action. Thoth reads telemetry before every check: the reading appeared at 958.7~s, and the plan changed at 958.8~s. The second is the reversible branch of \eqref{eq:adm}. Opening the shutter, waiting and resetting are reversible actions, so they were admitted without the ground. Earth had set at 925~s, and the run contains no ground request. The planner opened the shutter and waited 180~s for the unit to cool, and the breaker closed again at 1521.2~s. Thoth refused 3 of the 57 actions that the planner proposed in this run. When no irreversible action is pending, the admissibility rule rarely intervenes.

The baselines show what pre-validated procedures cannot do. The original PLEXIL and \EP{} procedures completed none of the seven contingency and sequential-fault missions, and each of these runs ended with a request for a new procedure and a hold (\cref{tab:coverage}). After revision, PLEXIL completed 4 and \EP{} 6 of the six contingency missions, at 38.9 and 224.3~min of engineering time and 2 and 14 corrected errors (\cref{fig:timeline}(b)). The declared model of \AKR{} was the same in all sixteen missions. \AKR{} separates what must be verified from what must adapt: the propositions, action classes and evidence rules are fixed, and the response to a new situation comes from the planner. The two revised PLEXIL runs that still failed show the difference. After the arm missed the BESS reset button, the procedure repeated the identical press, whereas the foundation model changed the parking distance or stowed the arm and pressed again. Ground contact follows the same division. \AKR{} sent a ground request in 3 of its 48 base runs, each an authorization that the mission requires, and the original baselines requested a new procedure in 7 and 8 of their 16 runs.

\begin{table}[t]
\caption{Coverage, mission time and effort of \AKR{} and the two baselines. An arrow separates the original from the revised baseline.}
\label{tab:coverage}
\centering
\footnotesize
\setlength{\tabcolsep}{3pt}
\begin{tabularx}{\columnwidth}{@{}Yccc@{}}
\toprule
 & \AKR{} & PLEXIL & \makecell{EUROPA$\rightarrow$\\PLEXIL} \\
\midrule
\multicolumn{4}{@{}l}{\textit{Evaluator successes over runs}} \\
\subrow Reference (7 missions) & 19/21 & 7/7 & 6/7 \\
\subrow Contingency (6) & 17/18 & 0/6$\,\to\,$4/6 & 0/6$\,\to\,$6/6 \\
\subrow Sequential fault (1) & 3/3 & 0/1 & 0/1 \\
\subrow Wrong record (2) & 5/6 & 2/2 & 2/2 \\
\midrule
\multicolumn{4}{@{}l}{\textit{Median time of successful runs (s)}} \\
\subrow Contingency & 790.4 & n/a$\,\to\,$762.0 & n/a$\,\to\,$776.0 \\
\subrow Reference repairs (4) & 263.2 & 250.5 & 230.0 \\
\subrow Reference deliveries (3) & 1201.0 & 464.0 & 471.0 \\
\midrule
\multicolumn{4}{@{}l}{\textit{Ground contact and engineering effort}} \\
\subrow Runs with a ground request (base, original) & 3/48 & 7/16 & 8/16 \\
\subrow Revision time, 4 classes (min) & 0 & 38.9 & 224.3 \\
\subrow Errors corrected in revision & 0 & 2 & 14 \\
\bottomrule
\end{tabularx}
\end{table}

On the missions that the procedures cover, the cost of \AKR{} is close to the baselines for repairs and higher for deliveries. The median successful reference repair took 263.2~s under \AKR{}, 250.5~s under PLEXIL and 230.0~s under \EP{}. The median reference delivery took 1201.0~s, 464~s and 471~s, and the median successful \AKR{} delivery used 29 skills against 9 in every baseline delivery. On the six contingency missions the median successful \AKR{} run took 790.4~s, against 762.0~s and 776.0~s for the revised baselines.

\Cref{cor:hold} states that a hold occurs only when no sound knowledge-based policy could execute the blocked action, so a hold is immediate. With Earth setting at 925~s, the latest start of a ground request is 616.5~s. Two runs sent the request before that time, received the grant and completed the repair. In the third the planner proposed the request at 723.5~s, and the run entered a certified hold at 723.6~s without sending a request whose reply would arrive after Earth set. In the three runs in which the ground refused, the hold began when the refusal arrived, at 1018.2~s, 960.5~s and 953.6~s, with the module still in its bay. In two \FFR{} runs under the same refusal the module had been removed before the refusal arrived.

\subsection{Evidence from physical measurements}
\label{sec:res-evidence}

\Cref{thm:sound} rests on \cref{asm:evidence}, and the full-stack runs measure the departure from it (\cref{tab:evidence-stats}). Among the conclusive judgments on 2177 actions, 0.26\% disagree with simulator truth; this is the selective risk of~\cite{geifman2017selective}. Conclusive judgments of pick, removal, press, stow and arm-ready actions never disagree. Of 5537 sensed values, 40 disagree, which gives $\eta=0.0072$. The two testbeds combine at this number. The Monte Carlo trials show how sensitive the conclusion of \cref{thm:sound} is to $\eta$, and the rover shows how large $\eta$ is: at 0.0072 the sweep of \cref{sec:res-sound} contains no hazardous action (\cref{fig:safety}(c)), and no \AKR{} run on the rover contains one. The rate differs by source, from 0.57\% for serial-plate reads to 4.69\% for the held-item reading.

\begin{table}[t]
\caption{Physical evidence over the 121 runs of the three acting rules: (a) effect realization and judgments per actuation skill, navigation excluded, and (b) sensed values per source. Conclusive judgments are counted per effect judgment, and the last row of (a) pools the 51,654 judgments of all skills, 37,990 of them from docking at a feature.}
\label{tab:evidence-stats}
\centering
\footnotesize
\setlength{\tabcolsep}{3pt}
\begin{tabularx}{\columnwidth}{@{}Ycccc@{}}
\toprule
(a) Actuation skill & Executions & \makecell{Full\\effect} & \makecell{Conclusive\\judgments} & \makecell{Judgment\\error (\%)} \\
\midrule
Dock at feature & 1003 & 0.879 & 0.989 & 0.29 \\
Dock to pose & 282 & 0.943 & 0.940 & 0.15 \\
Pick & 259 & 0.710 & 0.675 & 0.00 \\
Place & 153 & 0.935 & 0.776 & 0.63 \\
Handover & 44 & 0.750 & 0.773 & 0.74 \\
Removal & 14 & 0.857 & 0.413 & 0.00 \\
Press & 172 & 0.517 & 0.593 & 0.00 \\
Stow & 194 & 0.928 & 1.000 & 0.00 \\
Arm ready & 56 & 1.000 & 0.946 & 0.00 \\
All judged actions & 2177 & 0.847 & 0.962 & 0.26 \\
\midrule
(b) Sensing source & \multicolumn{2}{c}{Sensed values} & \multicolumn{2}{c}{Disagreeing with truth} \\
\midrule
Serial-plate read & \multicolumn{2}{c}{4709} & \multicolumn{2}{c}{27 (0.57\%)} \\
Localization & \multicolumn{2}{c}{684} & \multicolumn{2}{c}{9 (1.32\%)} \\
Held-item reading & \multicolumn{2}{c}{64} & \multicolumn{2}{c}{3 (4.69\%)} \\
Arm state & \multicolumn{2}{c}{64} & \multicolumn{2}{c}{0 (0.00\%)} \\
Camera question & \multicolumn{2}{c}{14} & \multicolumn{2}{c}{0 (0.00\%)} \\
Held-item check & \multicolumn{2}{c}{2} & \multicolumn{2}{c}{1 (50.0\%)} \\
All sensed values & \multicolumn{2}{c}{5537} & \multicolumn{2}{c}{40 (0.72\%)} \\
\bottomrule
\end{tabularx}
\end{table}

Three-valued evidence trades coverage for correctness. The evidence rules abstain when a measurement is inconclusive. The share of conclusive judgments is 0.413 for removal, 0.593 for press and 0.675 for pick, and 22.4\% of placement judgments are inconclusive, mostly because the destination lies outside the settled camera view. The press value follows from its rule, which leaves the equipment effect to telemetry. In exchange, a conclusive judgment is almost never wrong. The algorithm carries the cost of abstention, and safety does not: the proposition stays undecided, and a later sensing action or episode decides it. Over all judged actions, 84.7\% of executions realized all their effects and 96.2\% of judgments were conclusive. These pooled rates are the parameters of the episode trials in \cref{sec:res-recovery}.

% ======================================================================
\section{Discussion}
\label{sec:discussion}

Safety depends on \cref{asm:evidence} and on no property of the planner. Assurance therefore improves with the fidelity of physical evidence, for example through multi-view verification and independent hardware interlocks on $\Airr$, whatever the architecture or scale of the planning model. The same division holds for cost and termination. \Cref{thm:cost} fixes the sensing at each decision point, and \cref{sec:res-sensing} reports the mission time that results. The skill budget, the planner-call cap and the certified hold bound a run under an incomplete planner and leave \cref{thm:sound} intact.

On a local radiation-tolerant processor or one in cis-lunar space, the admissibility check and the deadline check $t\le w_\sigma$ read the mission clock at dispatch, so \cref{thm:sound} holds for any inference latency. Latency determines how many episodes fit into a time window, such as the $30$~s before a thermal re-trip or the interval before $T_{\mathrm{vis}}-c_g$. 

The evaluation fixes the solar elevation at $12^{\circ}$, draws observation errors independently in the Monte Carlo trials, and uses one planning model. The next steps are low-angle polar shadowing, correlated visual occlusions, several planning models on a diagnostic benchmark, and a skill that recovers an item from the regolith.

% ======================================================================
\section{Conclusion}
\label{sec:conclusion}

Lunar-base maintenance and logistics have been formulated as safety-constrained contingent planning with potentially erroneous records, three-valued physical evidence, irreversible actions, and costly deadline-constrained sensing. Asymmetric knowledge-level replanning executes plans synthesized by an untrusted foundation model. It permits one unverified attempt per reversible precondition and admits an irreversible action only on verified knowledge, after a minimum-cost sensing cover. Under sound physical evidence, no irreversible action executes on a false precondition for any planner or record; the sensing before an irreversible action is bounded from below for every sound knowledge-based policy; and execution with a complete planner ends with the goal verified or in a certified hold. \AKR{} executed no hazardous action in the 21,000 main-grid Monte Carlo trials with sound observations or in its 58 full-stack rover runs. Its median run took 1.10 times the fastest run on the same mission, against 1.50 for the \KR{}. It completed 20 of 21 contingency and sequential-fault runs with an unchanged task model, where the original PLEXIL and \EP{} procedures completed none. Assurance rests on a compact admissibility rule and on physical evidence that can be qualified before flight.

\FloatBarrier
\bibliographystyle{cas-model2-names}
\bibliography{refs}

\end{document}